\documentclass[lettersize,journal]{IEEEtran}
\IEEEoverridecommandlockouts

\usepackage{msc}
\usepackage{algorithmic}
\usepackage{graphicx}
\usepackage{acronym, comment, subcaption}
\usepackage{tabularx, pifont}
\usepackage[ruled,vlined,linesnumbered]{algorithm2e}
\usepackage{hyperref} 
\usepackage{mathtools}
\usepackage[colorinlistoftodos,textsize=small]{todonotes}
\usepackage{pdflscape}
\usepackage{lipsum}
\usepackage{booktabs}
\usepackage{multirow}
\usepackage{balance}
\usepackage{tikz}
\usetikzlibrary{chains,shapes.multipart}
\usetikzlibrary{shapes,calc}
\usetikzlibrary{automata,positioning}

\newcommand{\cmark}{\ding{51}}
\newcommand{\xmark}{\ding{55}}

\SetCommentSty{mycommfont}
\newcommand\revised[1]{\textcolor{black}{#1}}

\newcolumntype{P}[1]{>{\centering\arraybackslash}p{#1}}

\graphicspath{{Figures/}}

\acrodef{IoT}{Internet of Things}
\acrodef{RF}{Radio Frequency}
\acrodef{RFF}{\ac{RF} Fingerprinting}
\acrodef{PHY}{Physical}
\acrodef{NN}{Neural Network}
\acrodef{DL}{Deep Learning}
\acrodef{ML}{Machine Learning}
\acrodef{AI}{Artificial Intelligence}
\acrodef{TinyML}{Tiny Machine Learning}
\acrodef{AP}{Access Point}
\acrodef{SDR}{Software Defined Radio}
\acrodef{RNN}{Recurrent Neural Network}
\acrodef{CNN}{Convolutional Neural Network}
\acrodef{ReLU}{Rectified Linear Unit}
\acrodef{FCDD}{Fully Convolutional Data Description}
\acrodef{SNR}{Signal-to-Noise Ratio}
\acrodef{SVM}{Support Vector Machine}
\acrodef{TFLite}{TensorFlow Lite}
\acrodef{IoD}{Internet of Drones}
\acrodef{IoV}{Internet of Vehicles}
\acrodef{IoMT}{Internet of Medical Things}
\acrodef{NLP}{Natural Language Processing}

\usepackage{scalerel}
\usepackage{tikz}
\usetikzlibrary{svg.path}

\definecolor{orcidlogocol}{HTML}{A6CE39}
\tikzset{
  orcidlogo/.pic={
    \fill[orcidlogocol] svg{M256,128c0,70.7-57.3,128-128,128C57.3,256,0,198.7,0,128C0,57.3,57.3,0,128,0C198.7,0,256,57.3,256,128z};
    \fill[white] svg{M86.3,186.2H70.9V79.1h15.4v48.4V186.2z}
                 svg{M108.9,79.1h41.6c39.6,0,57,28.3,57,53.6c0,27.5-21.5,53.6-56.8,53.6h-41.8V79.1z M124.3,172.4h24.5c34.9,0,42.9-26.5,42.9-39.7c0-21.5-13.7-39.7-43.7-39.7h-23.7V172.4z}
                 svg{M88.7,56.8c0,5.5-4.5,10.1-10.1,10.1c-5.6,0-10.1-4.6-10.1-10.1c0-5.6,4.5-10.1,10.1-10.1C84.2,46.7,88.7,51.3,88.7,56.8z};
  }
}

\newcommand\orcidicon[1]{\href{https://orcid.org/#1}{\mbox{\scalerel*{
\begin{tikzpicture}[yscale=-1,transform shape]
\pic{orcidlogo};
\end{tikzpicture}
}{|}}}}
\usepackage{hyperref}

\graphicspath{{Figures/}}

\begin{document}

\title{Edge AI-based Radio Frequency Fingerprinting for IoT Networks}

\author{
	\IEEEauthorblockN{
        Ahmed Mohamed Hussain~\orcidicon{0000-0003-4732-9543}\IEEEauthorrefmark{1},~\IEEEmembership{Member,~IEEE,} Nada Abughanam~\orcidicon{0000-0002-0301-6724}\IEEEauthorrefmark{2}, and Panos Papadimitratos~\orcidicon{0000-0002-3267-5374}\IEEEauthorrefmark{1},~\IEEEmembership{Fellow,~IEEE}
 }
  \\
	\IEEEauthorblockA{ \IEEEauthorrefmark{1}Networked Systems Security (NSS) Group --
    KTH Royal Institute of Technology, Stockholm, Sweden \\ \IEEEauthorrefmark{2}Electrical Engineering Department -- Qatar University, Doha, Qatar\\ ahmed.hussain@ieee.org, nada.abughanam@qu.edu.qa, papadim@kth.se}

\thanks{This paper was produced by the IEEE Publication Technology Group. They are in Piscataway, NJ.}
\thanks{Manuscript received December~2024.}}

\markboth{Journal of \LaTeX\ Class Files,~Vol.~XX, No.~XX, December~2024}%
{Shell \MakeLowercase{\textit{A. Hussain et al.}}: A Sample Article Using IEEEtran.cls for IEEE Journals}


\maketitle

\begin{abstract}
The deployment of the Internet of Things (IoT) in smart cities and critical infrastructure has enhanced connectivity and real-time data exchange but introduced significant security challenges. While effective, cryptography can often be resource-intensive for small-footprint resource-constrained (i.e., IoT) devices. Radio Frequency Fingerprinting (RFF) offers a promising authentication alternative by using unique RF signal characteristics for device identification at the Physical (PHY)-layer, without resorting to cryptographic solutions. The challenge is two-fold: how to deploy such RFF in a large scale and for resource-constrained environments. Edge computing, processing data closer to its source, i.e., the wireless device, enables faster decision-making, reducing reliance on centralized cloud servers. Considering a modest edge device, we introduce two truly lightweight Edge AI-based RFF schemes tailored for resource-constrained devices. We implement two Deep Learning models, namely a Convolution Neural Network and a Transformer-Encoder, to extract complex features from the IQ samples, forming device-specific RF fingerprints. We convert the models to TensorFlow Lite and evaluate them on a Raspberry Pi, demonstrating the practicality of Edge deployment. Evaluations demonstrate the Transformer-Encoder outperforms the CNN in identifying unique transmitter features, achieving high accuracy ($>$ 0.95) and ROC-AUC scores ($>$ 0.90) while maintaining a compact model size of 73KB, appropriate for resource-constrained devices.
\end{abstract}


%
\section{Introduction}
\label{sec:intro}
\IEEEPARstart{T}{he} growth of the \acf{IoT}, supported by the evolution of networks, from 4G to 5G and forward, has fundamentally transformed various sectors by enabling pervasive connectivity and real-time data exchange among heterogeneous devices. \ac{IoT} devices are deployed at the network Edge and are often utilized for smart home, healthcare, industrial automation, and smart city applications~\cite{al2015internet}. However, the inherent resource constraints and massive scale deployment present significant security challenges, particularly in ensuring device-generated data legitimacy, authenticity, and integrity. 

Security in this landscape is critical, as the vast number of interconnected devices increases the attack surface, leading to networks being more vulnerable to attacks such as breaches, data tampering, and unauthorized access. A key challenge is to be practical for small-footprint resource-constrained devices. Cryptographic protocols can be of central importance, especially in terms of device authentication, even though the management of cryptographic keys can be complex. Further, cryptography can be hard, if not impossible, to implement in some classes of \ac{IoT} devices - notably, backscatter tags~\cite{narayanan2021harvestprint, 9170604}. Hence, \ac{RFF} has been proposed. Moreover, resources at the edge, with edge devices more powerful than IoT devices, can be used to implement security mechanisms. 

The challenge lies in efficiently deploying effective \ac{RFF} in resource-constrained environments. \ac{RFF} is a technique that uses the \ac{PHY}-layer signal characteristics for wireless device authentication~\cite{soltanieh2020review}. The imperfections in \ac{RF} transmitters create a unique ``fingerprint'' for each device. These fingerprints are based on the inherent variations in the transmitted signals and can be captured through the In-phase (I) and Quadrature (Q) (IQ) components. This allows for reliable device identification and authentication without relying on traditional cryptographic approaches~\cite{zhu2024secure}. \ac{RFF} enables differentiating between devices, even when they operate on the same protocol and frequency band.

To enhance \ac{RFF}, \ac{AI} has been introduced to extract and learn complex patterns from \ac{RF} signals. \ac{AI}-based \ac{RFF} leverages \ac{ML}, specifically \ac{DL}, to improve feature extraction that other methods may lack, hence improving the identification accuracy~\cite{AlHazbiAHSSGOPP:C:2024}. However, the fundamental question is \textit{whether edge devices can support such operations or not}. While edge devices are typically more powerful than low-end \ac{IoT} devices, they still lack the ample computational resources found in centralized cloud servers, making their ability to efficiently run \ac{AI}-based \ac{RFF} infeasible.

For instance, devices such as Raspberry Pi or NVIDIA Jetson, while more capable than basic \ac{IoT} devices are still limited in terms of processing power, memory, and energy consumption. At the same time, \ac{AI}-based \ac{RFF} is not computationally cheap; building and training \ac{DL} architectures typically demands significant resources. Additionally, real-time processing and inference that ensures accurate identification/authentication requires sufficient resources that edge devices may lack. The challenge lies in developing lightweight, optimized \ac{ML} models that maintain high performance while being computationally feasible for edge devices.

A limited number of recent contributions~\cite{jian2021radio, wu2023radio} discussed deploying and operating \revised{\ac{RFF}-based authentication} on the edge. Only~\cite{jian2021radio}  considers \ac{AI}-based \ac{RFF} at the edge: The authors rely on transfer learning (i.e., use models architectures such as ResNet50~\cite{he2016deep} that are not tailored for \ac{RFF}) to build the \acf{DL} network and use pruning to reduce the model size. Additionally, they do not consider evaluating other \ac{DL} models or introducing lightweight architectures. Indeed, deploying \ac{AI}-based \ac{RFF} models on the edge enables real-time device authentication, reducing latency and further strengthening the network's security, hence the need for easy-to-build and train lightweight, high-performing, and deployable models.

To bridge this gap, this paper presents two lightweight \ac{AI}-based \ac{RFF} implementations tailored for edge devices. Our primary focus is on the training and deployment of \ac{RFF} models tailored for edge/\ac{IoT} devices, ensuring robust and scalable identification/authentication using \ac{PHY}-layer characteristics. We dissect the building, training, and deployment of lightweight models on the edge, ensuring that they can perform a robust \ac{RFF}-based authentication without performance degradation.

\vspace{0.25cm}

\noindent\textbf{Contributions.} In this paper, we introduce and evaluate two truly lightweight \ac{DL} architectures specifically designed for \ac{PHY}-layer authentication on edge devices. Additionally, unlike the state-of-the-art~\cite{jian2021radio,wu2023radio}, the presented models do not rely on \textit{transfer learning} and ensure high training, testing, and prediction accuracy without the need for pre-trained models. We convert and optimize the \ac{DL} models using TensorFlow Lite to reduce the size and enhance inference speed, ensuring that the models are suitable for deployment on edge devices, even those with limited computational resources. Moreover, we present the first Transformer-based \ac{RFF} model that is deployable on edge devices. The implementation (code) will be released upon the acceptance of the paper. Finally, we evaluate the performance over 28 transmitters (from an open source dataset~\cite{hanna_wisig_2022}) and show that the Transformer encoder achieves and maintains high accuracy and ROC-AUC, outperforming traditional \ac{CNN}. 

\vspace{0.25cm}

\noindent{\bf Paper Organization.} Section~\ref{sec:preliminaries} provides an overview of the preliminaries. Section~\ref{sec:scenario_adversarial} details the considered system model and states the assumptions. Section~\ref{sec:methodology} discusses the implementation of our proposed framework, including the Deep Learning and Transformer Encoder architecture, as well as the TinyML conversion and optimization steps. Section~\ref{sec:performance_evaluation} presents a comprehensive performance evaluation, including detailed analysis and results. Section~\ref{sec:rw} presents the relevant related work and compares our contributions to the existing body of research. Finally, Section~\ref{sec:conclusions} concludes the paper with a summary of our findings and potential directions for future research.

\section{Preliminaries}
\label{sec:preliminaries}
This section discusses the concepts essential for understanding the rest of the paper, specifically \acl{DL}, Transformers, TinyML, and \acl{RFF}.

\subsection{Deep Learning, Transformers, and TinyML}
\textbf{\acf{DL}} is a subset of \acf{ML} characterized by using neural networks with several layers, capable of learning from large datasets to perform various tasks such as \ac{NLP} and speech recognition. Within this broader domain, \acfp{CNN} represent a specialized type of \ac{NN} that performs well in processing different types of data~\cite{alzubaidi2021review}. \acp{CNN} consists of convolutional layers, where parametrized filters, or kernels, slide over the input data to create feature maps by applying these filters to different input parts~\cite{o2015introduction}. Through this process, \acp{CNN} hierarchically learns and combines basic patterns such as edges, textures, and shapes, which increase in complexity through successive layers, making them highly effective for computer vision, image and video recognition, object detection, and image segmentation tasks.

\textbf{Transformer} is a \ac{NN} architecture that generates output data by analyzing and understanding the context of the input data~\cite{vaswani2017attention}. Unlike the sequential processing of data performed by \acp{RNN}, Transformers can process the entire data sequence simultaneously, allowing for extracting relationships and context within the data sequence. The overall Transformer architecture, \revised{(introduced in~\cite{vaswani2017attention})}, includes two primary components: the \textit{encoder} and \textit{decoder}. The encoder is tasked with processing the input data sequence, transforming it into a continuous representation that encapsulates contextual information. This is achieved through multiple layers, each incorporating a multi-head self-attention mechanism and a position-wise fully connected feed-forward network. The self-attention mechanism enables the encoder to assign varying degrees of importance to different parts of the input sequence, thereby generating a comprehensive representation of the data. Conversely, the decoder generates the output sequence by utilizing the representations produced by the encoder along with previously generated tokens. Similar to the encoder, the decoder is composed of multiple layers featuring a multi-head self-attention mechanism and a position-wise feed-forward network. Additionally, the decoder includes an encoder-decoder attention mechanism, which aligns the output sequence with relevant parts of the input sequence by attending to the encoder's outputs. This mechanism ensures that the decoder accurately maps the relationships between the input and output sequences.

\textbf{TinyML} facilitates the deployment of \ac{ML} and \ac{DL} models on small, resource-constrained, low-powered devices. Traditionally, \ac{DL} models are run on servers with powerful computational resources available, where edge devices are merely responsible for collecting data and transmitting it to the server for processing and classification. With the emergence of edge \ac{AI} and TinyML, \ac{DL} models can be deployed on the edge device itself, conserving transmission bandwidth. This is primarily important in smart cities, where areas such as smart healthcare~\cite{hayyolalam2021edge}, smart living~\cite{yan2022survey}, and smart mobility~\cite{paiva2021enabling} applications require the use of edge AI to facilitate many processes. This includes self-driving cars, autonomous robots, and AI-powered smartphones~\cite{singh2023edge}. Through the utilization of \ac{AI} on edge, the collection of the data and inference is performed almost instantly on the edge itself without the need to send the data to the server for processing. To achieve deployable and lightweight models, several optimization methods can be used to convert the \ac{DL} models and reduce their size in order to deploy them on low-power and low-memory edge devices without significantly affecting performance, including model quantization and pruning~\cite{liang2021pruning}.

\subsection{Radio Frequency Fingerprinting}
\acf{RFF} is a technique that exploits the inherent hardware imperfections in \acf{RF} devices to identify them based on their emitted signals. The main principle of \ac{RFF} is that there exist no two devices of the same make and model that emit the same \ac{RF} signals, due to hardware imperfections that lead to variations. These result in unique device-specific characteristics in the \ac{RF} signal~\cite{elmaghbubdistinguishable}, which can be used as a fingerprint for device identification/authentication. The development pipeline of \ac{RFF} systems constitutes three phases~\cite{AlHazbiAHSSGOPP:C:2024}: Signal Acquisition and Preprocessing, Model Training, and Deployment.

To train and build a \ac{DL} architecture for \ac{RFF}, signal acquisition is initially done in a controlled environment, where \ac{RF} signals emitted by wireless devices are captured and stored as IQ samples, which retain the characteristics of the device's transmitter. Following the acquisition, the data is preprocessed and prepared for model training. Preprocessing includes feature selection, data augmentation, normalization, and noise removal. Model training includes designing and training a \ac{DL} model that effectively captures the distinct patterns corresponding to the unique \ac{RF} fingerprints of wireless devices from the raw IQ samples. Finally, deploying the trained \ac{DL} model into real systems to identify transmitting wireless devices using incoming RF fingerprints with the transmitters used in the model's training. There exist several contributions, e.g.,~\cite{elmaghbubdistinguishable, 10288752}, on the characteristics and type of features that could be extracted from the signal and used to perform DL-based \ac{RFF}.

\begin{figure}[!h]
    \centering
    \includegraphics[width=0.7\columnwidth]{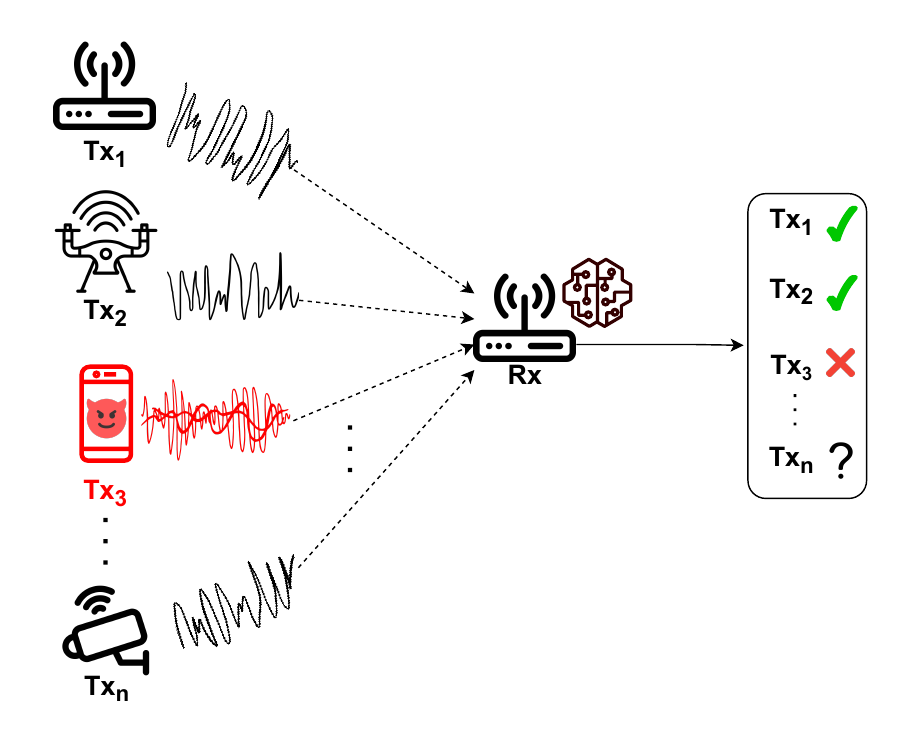}
    \caption{System model assumed in this paper. A set of \ac{IoT} devices with transmitters Tx$_{1, ...,\text{n}}$ are deployed in an attempt to connect to an edge device, i.e., an \ac{AP}, with receiver Rx. The \ac{AP} uses \ac{PHY}-layer information to identify/authenticate each \ac{RF} Fingerprint against a pre-trained model, allowing only authorized devices to connect to it.}
    \label{fig:scenario_adversarial}
\end{figure}

\begin{figure*}[!htpb]
    \centering
    \includegraphics[width=0.65\textwidth]{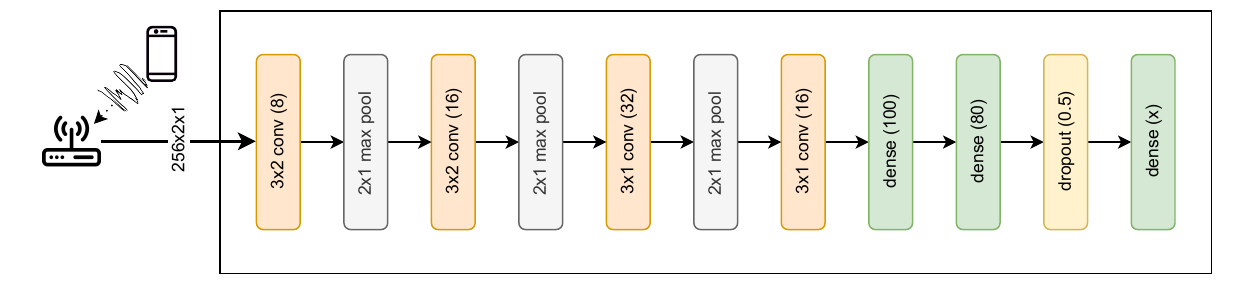}
    \caption{Structure and details of the implemented \ac{CNN}.}
    \label{fig:cnn_arch}
\end{figure*}

\begin{figure*}[!htpb]
    \centering
    \includegraphics[width=0.65\textwidth]{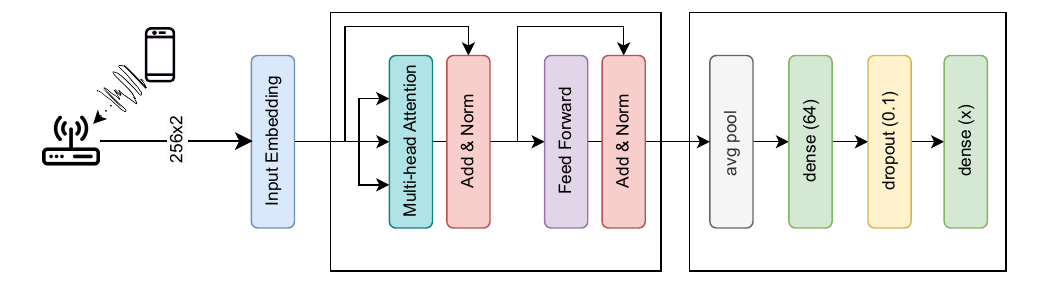}
    \caption{Structure and details of the implemented Transformer encoder.}
    \label{fig:transformer_arch}
\end{figure*}

\section{System Model and Problem Statement}
\label{sec:scenario_adversarial}

Figure~\ref{fig:scenario_adversarial} depicts the system model adopted in this paper. We consider a typical \ac{IoT} deployment where multiple devices, with transmitters, denoted $ \text{Tx}_1, \text{Tx}_2, \ldots, \text{Tx}_n$, attempt to \revised{connect} to an edge device, an \ac{AP}, that is equipped with a receiver Rx.  Upon receiving an \ac{RF} signal, the \ac{AP} processes the raw IQ samples to extract their distinctive features. These features are then evaluated against a pre-trained \ac{DL} model deployed on the \ac{AP}. This model, trained using \ac{RF} fingerprints from authorized devices, leverages the inherent \ac{PHY}-layer characteristics captured in the IQ samples to perform accurate device identification and authentication. An unauthorized device that is not part of the network (depicted in red) attempts to connect to the \ac{AP}. The \ac{AP} evaluates the \ac{RF} fingerprint of the unauthorized device against the pre-trained model. Since its \ac{RF} fingerprint will not match any known authorized devices, the \ac{AP} prevents/rejects the device from connecting.

We emphasize that introducing a new, unseen class during the developed model's test, evaluation, or inference phases constitutes an open-set problem, as discussed in~\cite{9040673} and further addressed in~\cite{WANG2024124537}. As noted in~\cite{WANG2024124537}, traditional \ac{RFF} approaches, including considerable \ac{ML}-based methods, struggle to generalize to unknown devices, as they are typically designed for closed-set scenarios where only known devices are present. The system operator, therefore, needs to ensure that the model is trained on a representative set of device data to handle real-world diversity.
The success of such an approach is heavily dependent on the model being trained with data from known system devices. This implies that the operator should regularly update the model to improve robustness and accuracy. 

As highlighted in~\cite{WANG2024124537}, addressing the challenge of unknown devices requires more sophisticated methods, such as contrastive learning, which creates discriminative representations of \acp{RFF} by generating positive and negative sample pairs. Exactly because the model relies on data from prior training, the open-set problem arises when unfamiliar classes are introduced. However, developing systems capable of handling open-set device classification is beyond the scope of this work, which assumes all classes are known during training.

We aim to deploy the trained models on edge devices. However, this is considered challenging due to constraints such as (i) the model size cannot be large as it affects the inference time and (ii) the fact that when the model's size is reduced, the trained model accuracy is decreased due to quantization that reduces the precision of the weights and activations in a model. Constraint (i) is presented in Section~\ref{sec:methodology}, while (ii) is presented in Section~\ref{sec:performance_evaluation}.


\section{Edge AI-based RFF}
\label{sec:methodology}
This section discusses the \acl{DL} and Transformer Encoder architecture implementation as well as the \ac{TinyML} conversion process. 

\subsection{Deep Learning}
Figure~\ref{fig:cnn_arch} illustrates the structure and details of the implemented \ac{CNN}. The implementation is done using TensorFlow~\cite{tf} and Keras~\cite{keras} libraries. The input layer reshapes the data to (256, 2, 1) to accommodate the 2D convolution operations. The model includes a series of convolutional and max-pooling layers. The first convolutional block employs 8 filters with a kernel size of (3, 2), followed by a max-pooling layer with a pool size of (2, 1). This is followed by the second convolutional block with 16 filters of the same kernel size and pooling configuration. The third block features 32 filters with a kernel size of (3, 1) and identical pooling. The fourth block reduces the filter count to 16, maintaining the kernel size (3, 1) without subsequent pooling to retain sufficient resolution for dense layers.

Post convolution, a flattening layer converts the 3D tensor output to a 1D vector. This vector is processed through two dense layers: (i) with 100 units and (ii) with 80 units, both employing \ac{ReLU} activation and L2 regularization with a factor of 0.0001 to mitigate overfitting. A dropout layer with a rate of 0.5 is also incorporated to prevent overfitting further. The output layer consists of a dense layer with units equal to the number of device classes, utilizing the softmax activation function and L2 regularization.

Finally, the model is compiled using the Adam optimizer~\cite{kingma2014adam, adam}, and sparse categorical cross-entropy as loss function. This lightweight \ac{CNN} architecture effectively captures spatial and temporal features from the input signals through convolutional and pooling layers, while the dense layers and dropout layers ensure robust feature learning and generalization. Table~\ref{tab:cnn_model} lists all the parameters and specifications for the developed \ac{CNN} model.

\subsection{Transformer Encoder}
Similar to the aforementioned \ac{CNN}, the presented Transformer encoder is implemented using TensorFlow and Keras libraries. The model architecture depicted in Figure~\ref{fig:transformer_arch} takes as input signals with a shape of (256, 2, 1). The input data is first projected into a higher-dimensional space with an embedding dimension of 64 using a dense layer. The model's core is a Transformer block, which includes a multi-head self-attention mechanism (2 heads, key dimension 64) and a feed-forward neural network with two dense layers (64 units each) and \ac{ReLU} activation. Layer normalization and dropout (rate 0.1) are applied for regularization. The output of the Transformer block is pooled and flattened to a 1D vector, followed by a dense layer with 64 units and \ac{ReLU} activation. Another dropout layer (rate 0.1) is applied before the final output layer, which uses the softmax activation function for multi-class classification, with the number of units equal to the number of device classes. The model is compiled using the Adam optimizer, sparse categorical cross-entropy loss, and accuracy as the evaluation metric. Table~\ref{tab:transformer_model} summarizes all the parameters and specifications of the implemented Transformer encoder.

\begin{table}[!h]
\centering
\caption{CNN model parameters summary}
\label{tab:cnn_model}
\resizebox{0.8\columnwidth}{!}{%
\begin{tabular}{@{}cc@{}}
\toprule
\textbf{Parameter} & \textbf{Specs} \\ \midrule
\textbf{Input Shape} & (256, 2, 1) \\
\textbf{Convolution Layers} & 4 layers (8, 16, 32, 16 filters respectively) \\
\textbf{Kernel Size} & (3, 2) and (3, 1) \\
\textbf{Pooling Layers} & 3 Max Pooling (pool size: (2, 1)) \\
\textbf{Dense Layers} & 3 layers (100, 80 units, 1 output layer) \\
\textbf{Dropout Rate} & 0.5 \\
\textbf{Output Activation} & Softmax \\
\textbf{Loss Function} & Sparse Categorical Crossentropy \\
\textbf{Optimizer} & Adam \\
\textbf{Activation Function} & \ac{ReLU} \\
\bottomrule
\end{tabular}%
}
\end{table}

\begin{table}[!h]
\centering
\caption{Transformer encoder model parameters summary}
\label{tab:transformer_model}
\resizebox{0.8\columnwidth}{!}{%
\begin{tabular}{@{}cc@{}}
\toprule
\textbf{Parameter} & \textbf{Specs} \\ \midrule
\textbf{Input Shape} &  (256, 2, 1) \\
\textbf{Embedding Dimension} & 64 \\
\textbf{Attention Heads} & 2 \\
\textbf{Feed-Forward Dimension} & 64 \\
\textbf{Dense Layers} & 3 layers (64, output layer) \\
\textbf{Dropout Rate} & 0.1 \\
\textbf{Output Activation} & Softmax \\
\textbf{Loss Function} & Sparse Categorical Crossentropy \\
\textbf{Optimizer} & Adam \\
\textbf{Activation Function} & \ac{ReLU} \\
\bottomrule
\end{tabular}%
}
\end{table}

\begin{figure*}[!htbp]
\centering
  \begin{subfigure}[b]{0.35\textwidth}
    \includegraphics[width=\columnwidth]{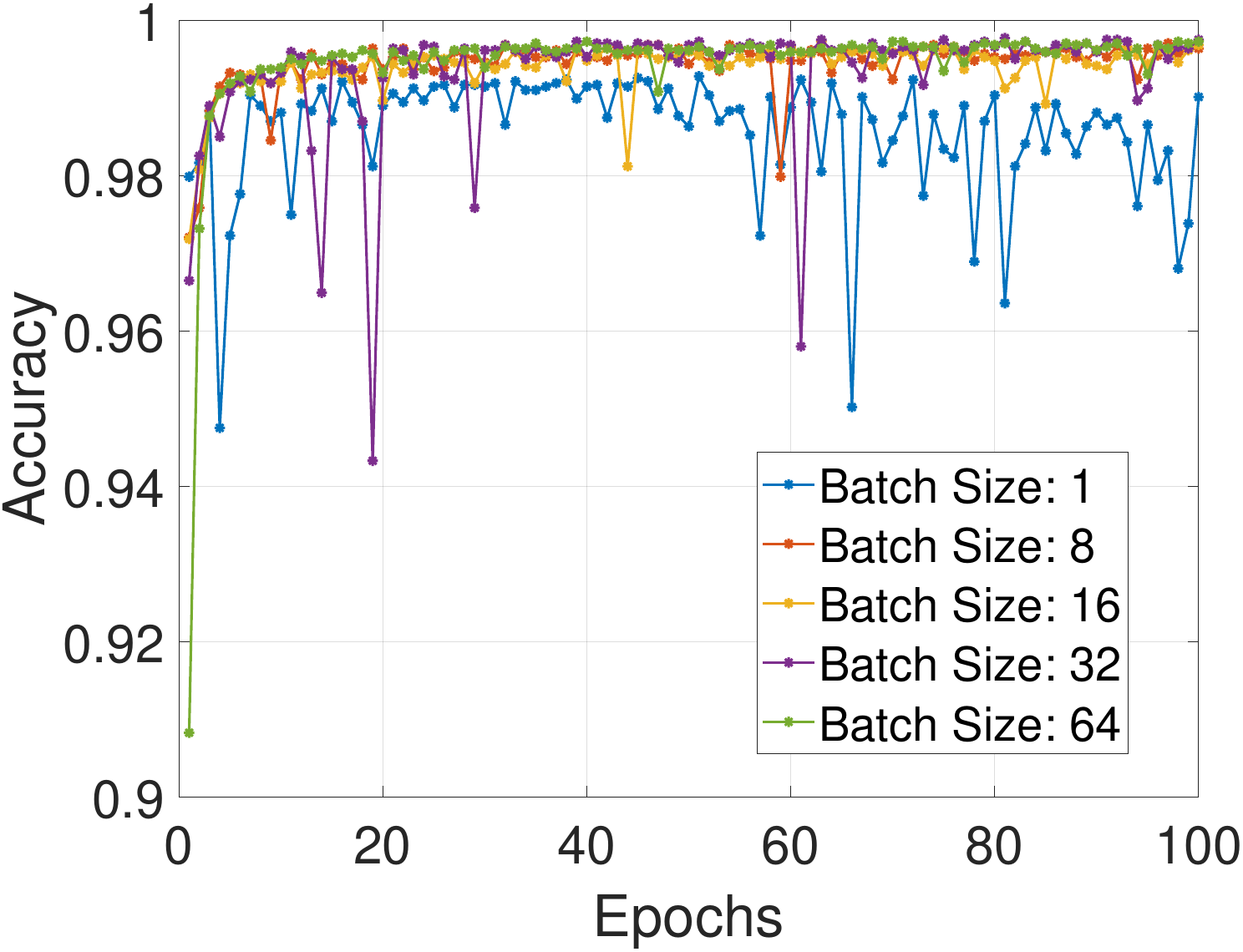}
    \caption{}
    \label{fig:cnn_validation_batchsize}
  \end{subfigure}
  \hspace{2cm}
   \begin{subfigure}[b]{0.35\textwidth}
    \includegraphics[width=\columnwidth]{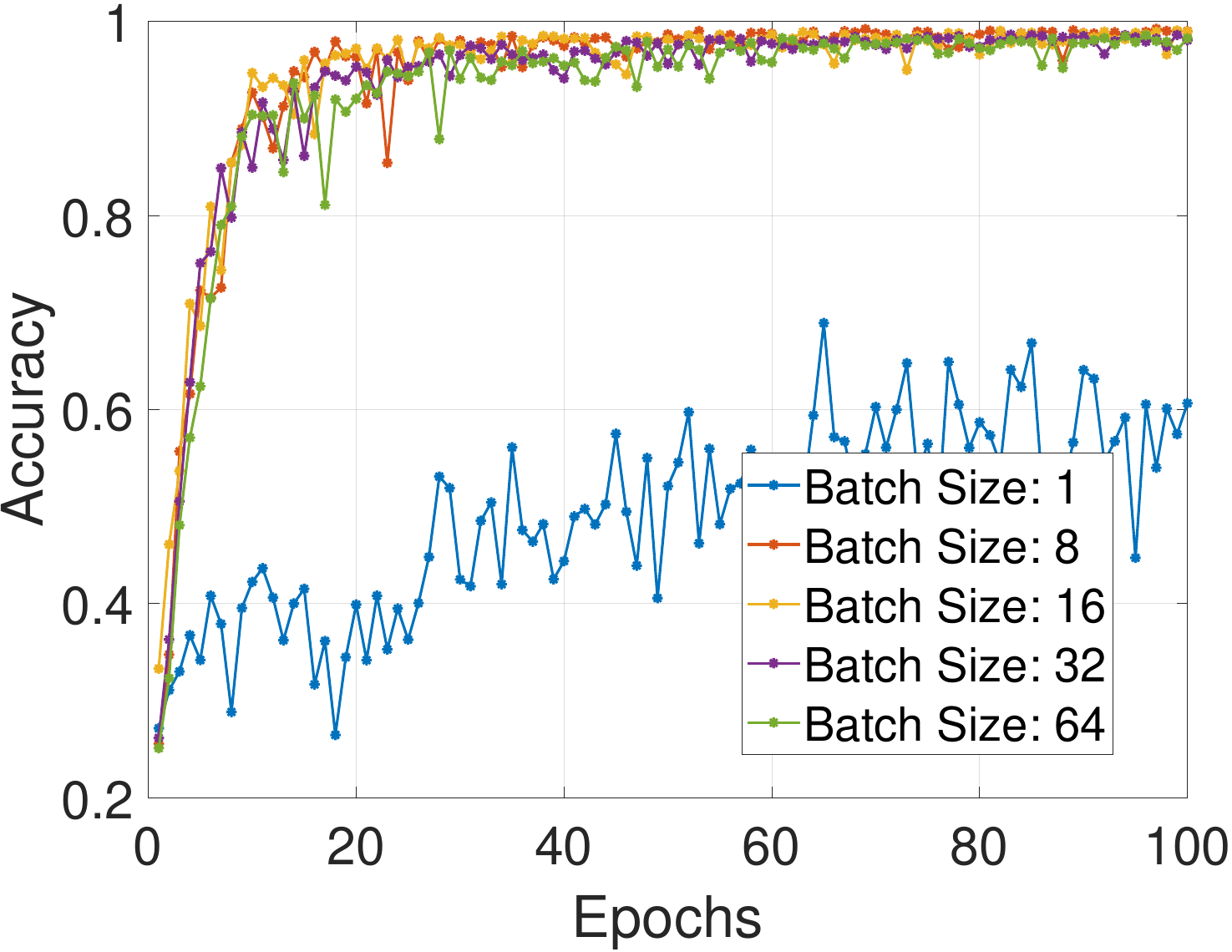}
    \caption{}
    \label{fig:transforemer_validation_batchsize}
  \end{subfigure}
    \caption{Batch sizes impact on model validation accuracy during model training for (a) CNN, and (b) Transformer}
    \label{fig:validation_batchsize}
\end{figure*}

\subsection{TinyML}
We convert the \ac{CNN} and Transformer encoder models using \ac{TFLite}. Several optimization techniques, including quantization, pruning, and clustering, can be applied to minimize model size and latency while maintaining precision. The steps below illustrate how to convert a regular \ac{CNN}~\cite{hussain2022jamming} or Transformer encoder model to a deployable and lightweight model on edge devices:
\begin{itemize}
    \item Initially, train and build the model using the regular Tensorflow.
    \item Specify the set of operations ``arget\_spec.supported\_ops'' that the converted TFLite model will support, namely, (i) \texttt{tf.lite.OpsSet.TFLITE\_BUILTINS}, and (ii) \texttt{tf.lite.O -psSet.SELE CT\_TF\_OPS}.
    \item Use the \textit{TFLiteConverter} to convert the model to a \ac{TFLite} format and invoke the convert function.
    \item Set the converter optimization attribute for post-training quantization to \texttt{tf.lite.Optimize.DEFAULT}. This enhances latency and reduces the model size~\cite{tflite_optimize}.
    \item Finally, deploy the converted ``.tflite'' model on the edge device and uses the TensorFlow Lite Interpreter to allocate tensors, provide input data, and perform inference.
\end{itemize}

\begin{table}
\centering
\caption{Model Sizes for Transformer and CNN Architectures}
\label{tab:conversion_size}
\resizebox{\columnwidth}{!}{%
\begin{tabular}{|c|c|c|}
\hline
\textbf{Type} & \textbf{Transformer (KB)} & \textbf{CNN (KB)} \\ \hline
Trained Model (Not converted) & 645.68 & 1437.17 \\\hline
TFLite & 210.2 & 462.02 \\\hline
TFLite Quantized & 73.27 & 123.8 \\ \hline
\end{tabular}%
}
\end{table}

The first operation (i) enables the use of \ac{TFLite}'s built-in operations, while the latter (ii) enables a selected set of TensorFlow operations that are necessary for certain models but are not part of \ac{TFLite} built-ins.

Tabel~\ref{tab:conversion_size} shows the sizes of all models: trained, \ac{TFLite}, and Quantized. The Transformer encoder model size is 645.68 KB. When the trained model is converted to a \ac{TFLite} model, the size is reduced to 210.20 KB, and when Quantization is applied, it is further reduced to 73.27 KB. This means that the \ac{TFLite} model is 3.07 times smaller, and the Quantized is 8.81 times smaller than the actual trained model. On the other hand, the \ac{CNN} model's size is 1437.17 KB, with its \ac{TFLite} reduced to 462.02 KB and the Quantized to 123.80 KB, resulting in the \ac{TFLite} model being 3.11 times smaller and the Quantized 11.61 times smaller than the original model.

\begin{figure*}[!htbp]
\centering
  \begin{subfigure}[b]{0.35\textwidth}
    \includegraphics[width=\columnwidth]{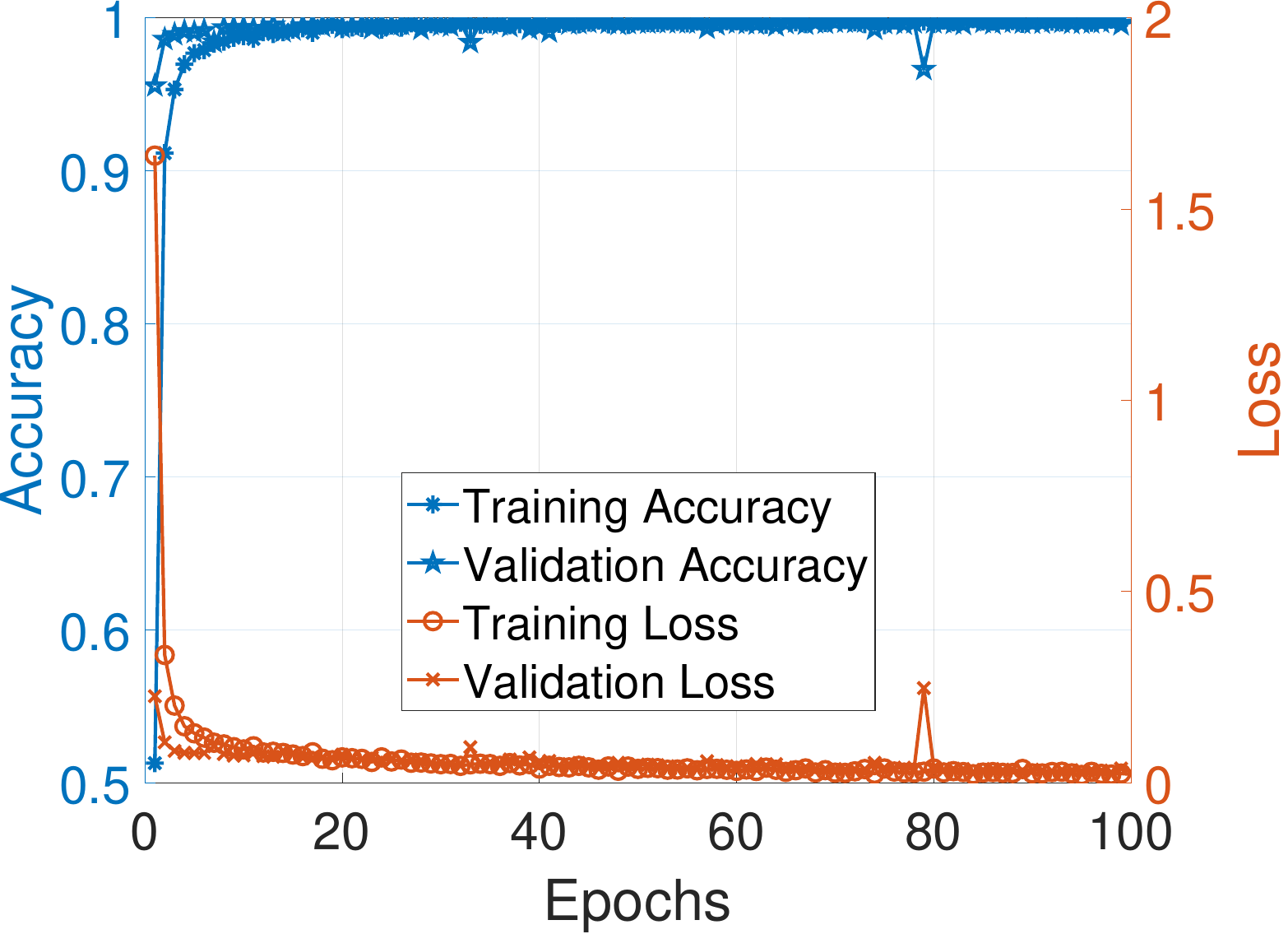}
    \caption{}
    \label{fig:cnn_train_validation}
  \end{subfigure}
  \hspace{2cm}
   \begin{subfigure}[b]{0.35\textwidth}
    \includegraphics[width=\columnwidth]{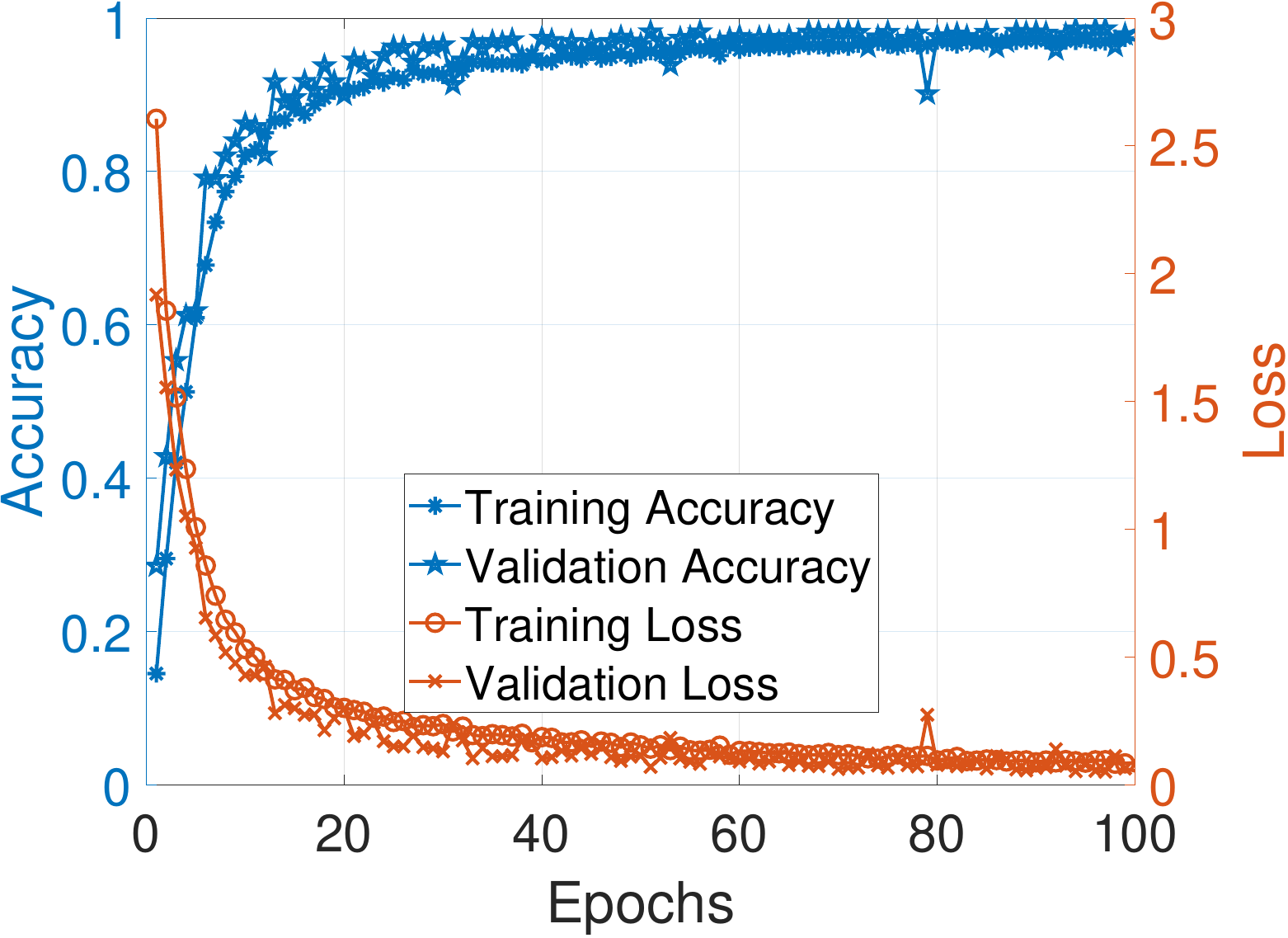}
    \caption{}
    \label{fig:transformar_train_validation}
  \end{subfigure}
    \caption{Model training and validation accuracy/loss as a function of number of epochs, for (a) \ac{CNN} and (b) Transformer encoder}
    \label{fig:tranning}
\end{figure*}

\section{Performance Evaluation}
\label{sec:performance_evaluation}
In this section, we discuss the performance of the \acl{CNN} and Transformer encoder models in terms of training, validation, and testing. Moreover, we evaluate the inference of the TinyML models on an edge device, namely, Raspberry Pi 4.

\subsection{Model Training and Evaluation}

\textbf{Training} utilizes the \textit{SingleDay} dataset provided by WiSig~\cite{hanna_wisig_2022} to train and build the presented \ac{DL} architectures. It consists of 800 WiFi signals generated by 28 transmitters collected over a one-day period and contains the IQ samples from signals emitted by each device. For a detailed description of the full dataset and how the data is collected, please refer to~\cite{hanna_wisig_2022}. The training is performed on a workstation with the following configurations: 12th Gen Intel(R) Core(TM) i9-12900K with a clock speed of 3.20 GHz, memory size of 64.0 GB, and NVIDIA RTX A4000 GPU with a memory size of 32 GB.

The models' training parameters are summarized in Table~\ref{table:training_options}. We considered the default learning rate, as it provides an acceptable performance and accuracy convergence for the \ac{DL} models. The number of epochs is 100, providing a reasonable balance between underfitting (insufficient learning) and overfitting (model learning noise instead of patterns). The training data is shuffled before each training epoch, preventing the model from learning patterns based on the order of training samples.

\begin{table}[!h]
    \centering
    \caption{Training parameters for \ac{DL} architectures.}
    \resizebox{0.7\columnwidth}{!}{%
    \begin{tabular}{cc}
        \toprule
            \textbf{Parameter} & \textbf{Value} \\
            \midrule
            Learning Rate & 0.001 (Default) \\
            Epochs & 100 \\
            Batch Size & 32 \\
            Shuffle & Every Epoch \\
            Validation Data & Random 20\% of the data \\
            \bottomrule
    \end{tabular}%
    }
    \label{table:training_options}
\end{table}

\textit{Data Preprocessing:} We preprocess the \ac{RF} signals to ensure uniform scaling across all samples to mitigate potential biases towards signals with higher magnitudes. This enhances the convergence and results in faster and more stable model learning. Furthermore, it prevents overflow and underflow issues arising from large variations in signal magnitudes. Each signal is scaled according to its inherent power.

\textit{Data Validation:} The training set is divided into training and validation sets with a ratio of 80/20. The chosen split ratio is common for the training and validation of \ac{DL} models. The validation set assesses the model's performance after each epoch during the tuning of the hyperparameters. 

\textit{Batch Size:} An important hyperparameter in training is batch size indicating the number of samples utilized in training the network in every epoch. A batch size of 32 is used to train both models, providing a good trade-off between stability (lower fluctuation in validation accuracy), as illustrated in Figure~\ref{fig:validation_batchsize}, and training time (faster than smaller batches but not too large to cause accuracy drops) as shown in Table~\ref{tab:batchsize_vs_time}.

\begin{figure*}[!htbp]
\centering
  \begin{subfigure}[b]{0.35\textwidth}
    \includegraphics[width=\columnwidth]{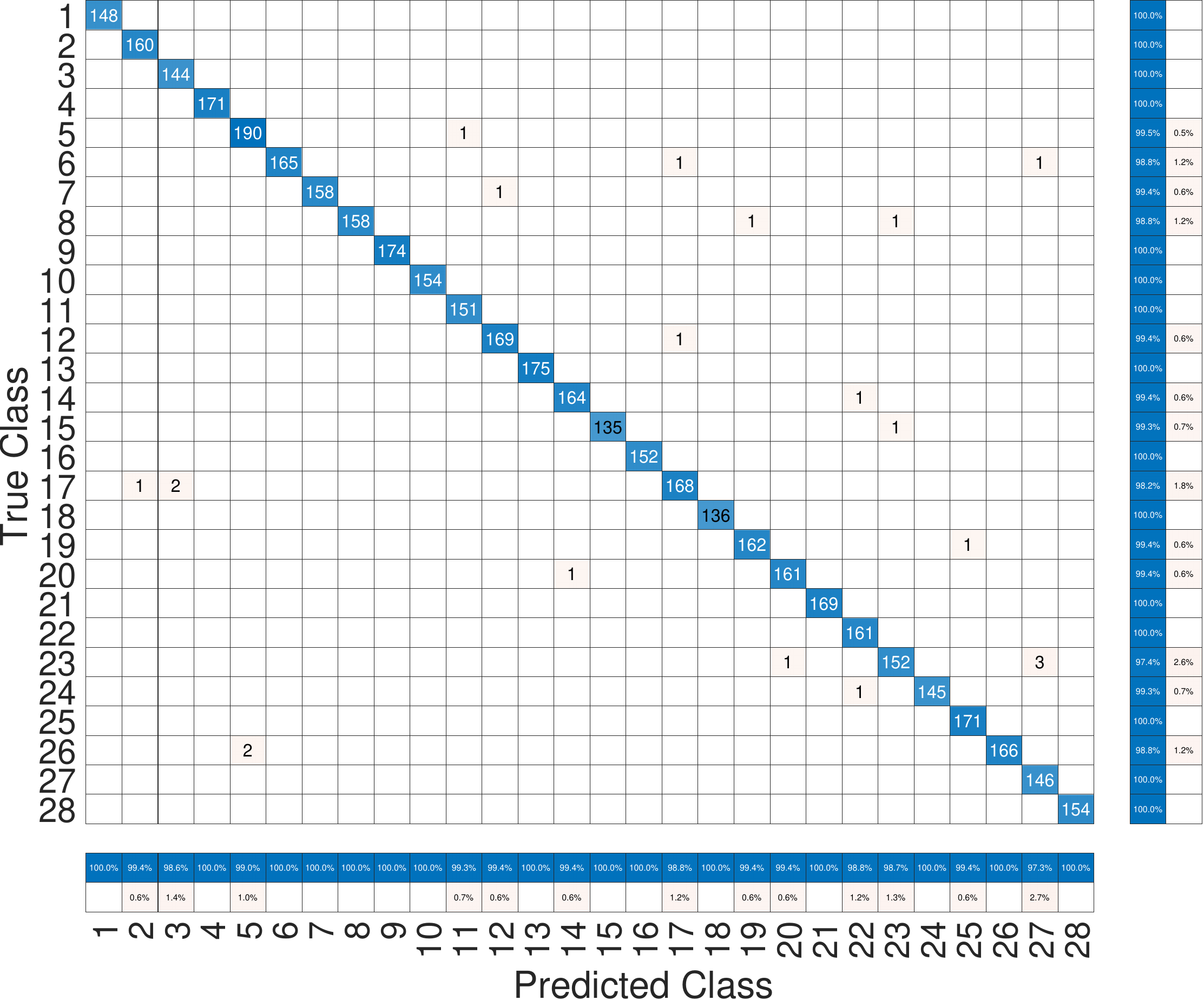}
    \caption{}
    \label{fig:confusion_matrix_keras_CNN}
  \end{subfigure}
  \hspace{2cm}
   \begin{subfigure}[b]{0.35\textwidth}
    \includegraphics[width=\columnwidth]{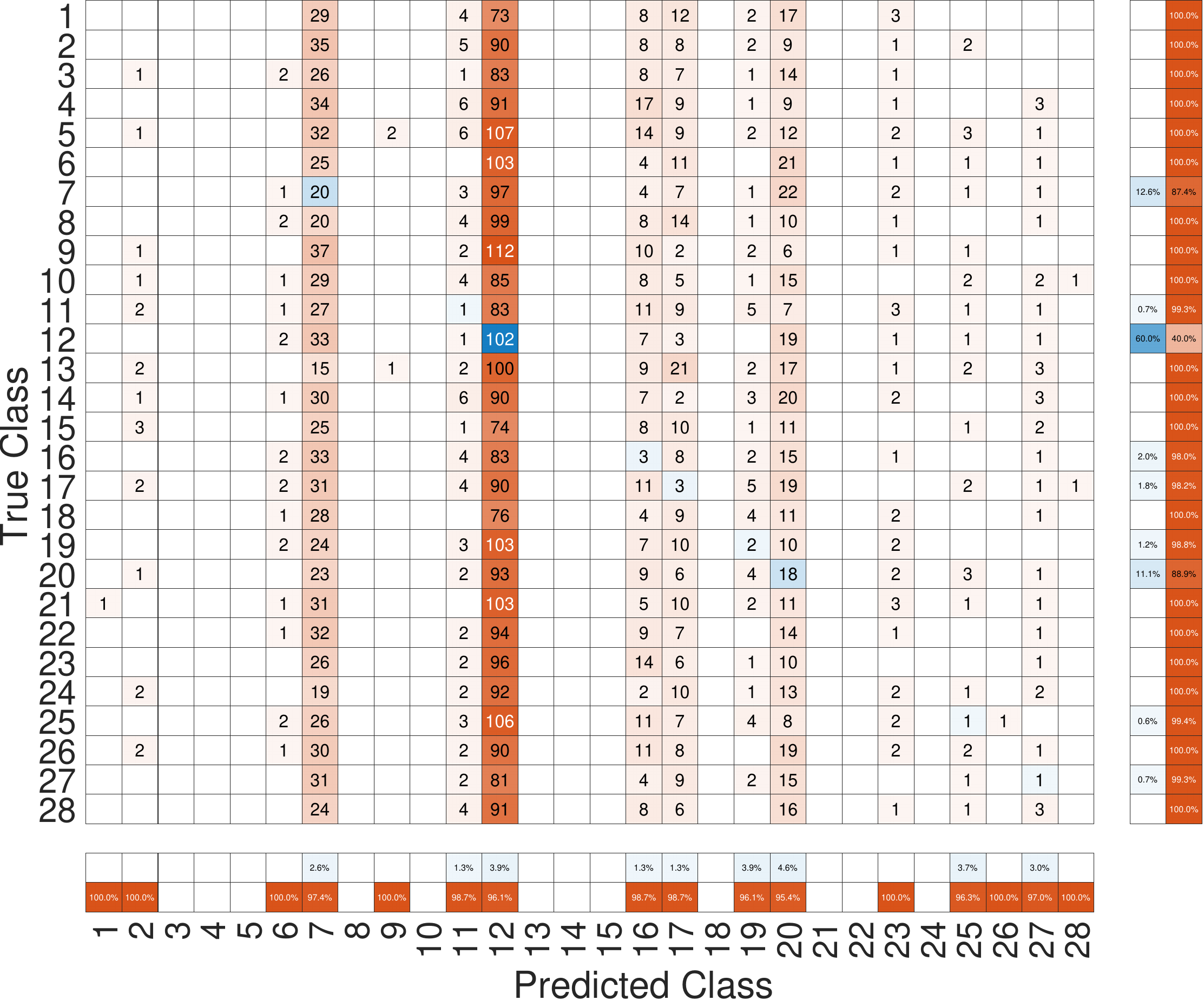}
    \caption{}
    \label{fig:confusion_matrix_keras_cnn_randomized}
  \end{subfigure}
    \caption{None converted \ac{CNN} model confusion matrix (a) without altering the IQ sequence (b) when randomizing the sequence.}
    \label{fig:confusion_matrix_keras_cnn}
\end{figure*}

\begin{figure*}[!htbp]
\centering
  \begin{subfigure}[b]{0.35\textwidth}
    \includegraphics[width=\columnwidth]{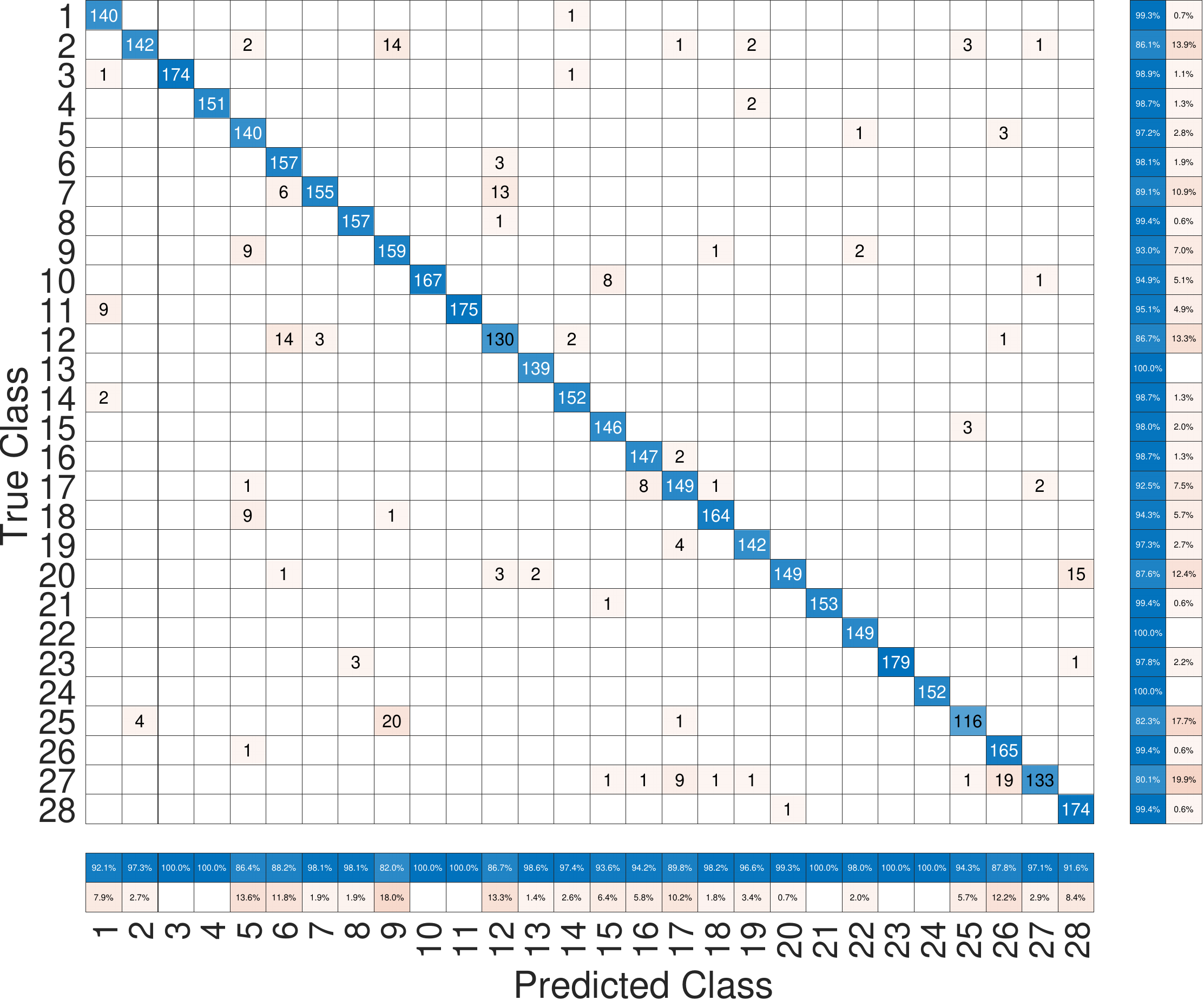}
    \caption{}
    \label{fig:confusion_matrix_keras_transformers}
  \end{subfigure}
\hspace{2cm}
   \begin{subfigure}[b]{0.35\textwidth}
    \includegraphics[width=\columnwidth]{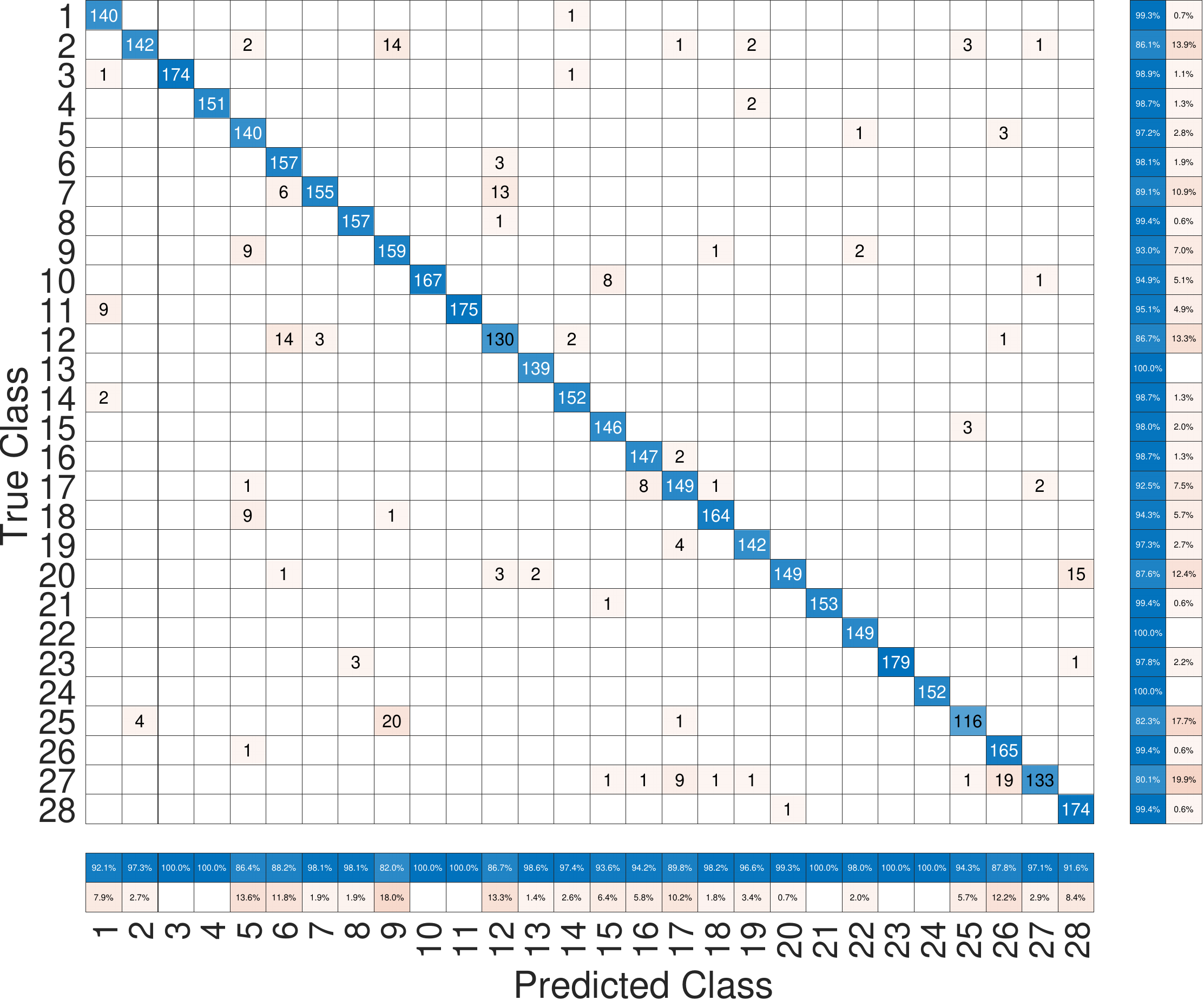}
    \caption{}
    \label{fig:confusion_matrix_keras_transformers_randomized}
  \end{subfigure}
    \caption{None converted Transformer encoder model confusion matrix (a) without altering the IQ sequence (b) when randomizing the sequence.}
    \label{fig:confusion_matrix_transformers}
\end{figure*}

\begin{figure*}[!htbp]
\centering
  \begin{subfigure}[b]{0.35\textwidth}
    \includegraphics[width=\columnwidth]{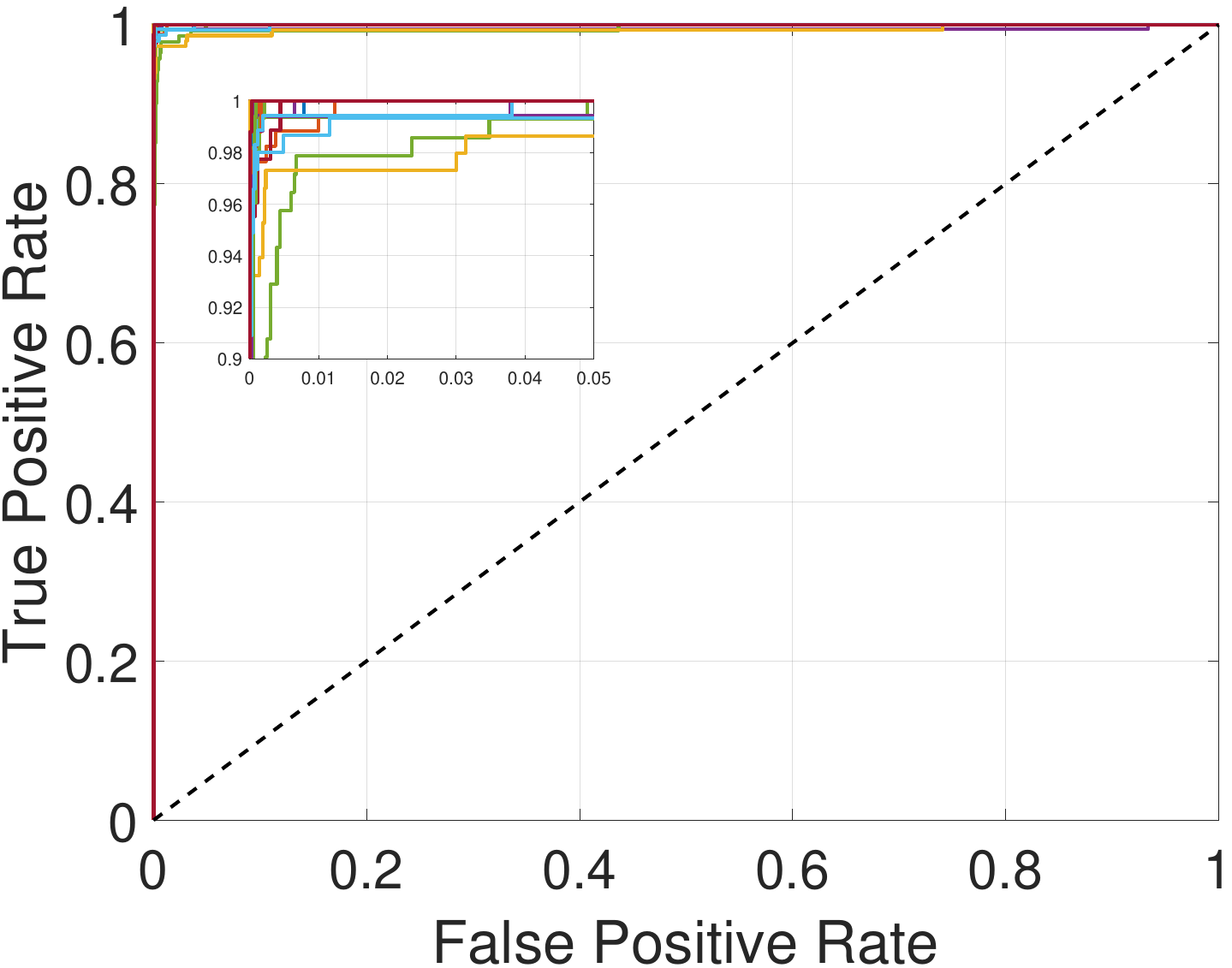}
    \caption{}
    \label{fig:roc_auc_transformar_tflite}
  \end{subfigure}
    \hspace{2cm}
   \begin{subfigure}[b]{0.35\textwidth}
    \includegraphics[width=\columnwidth]{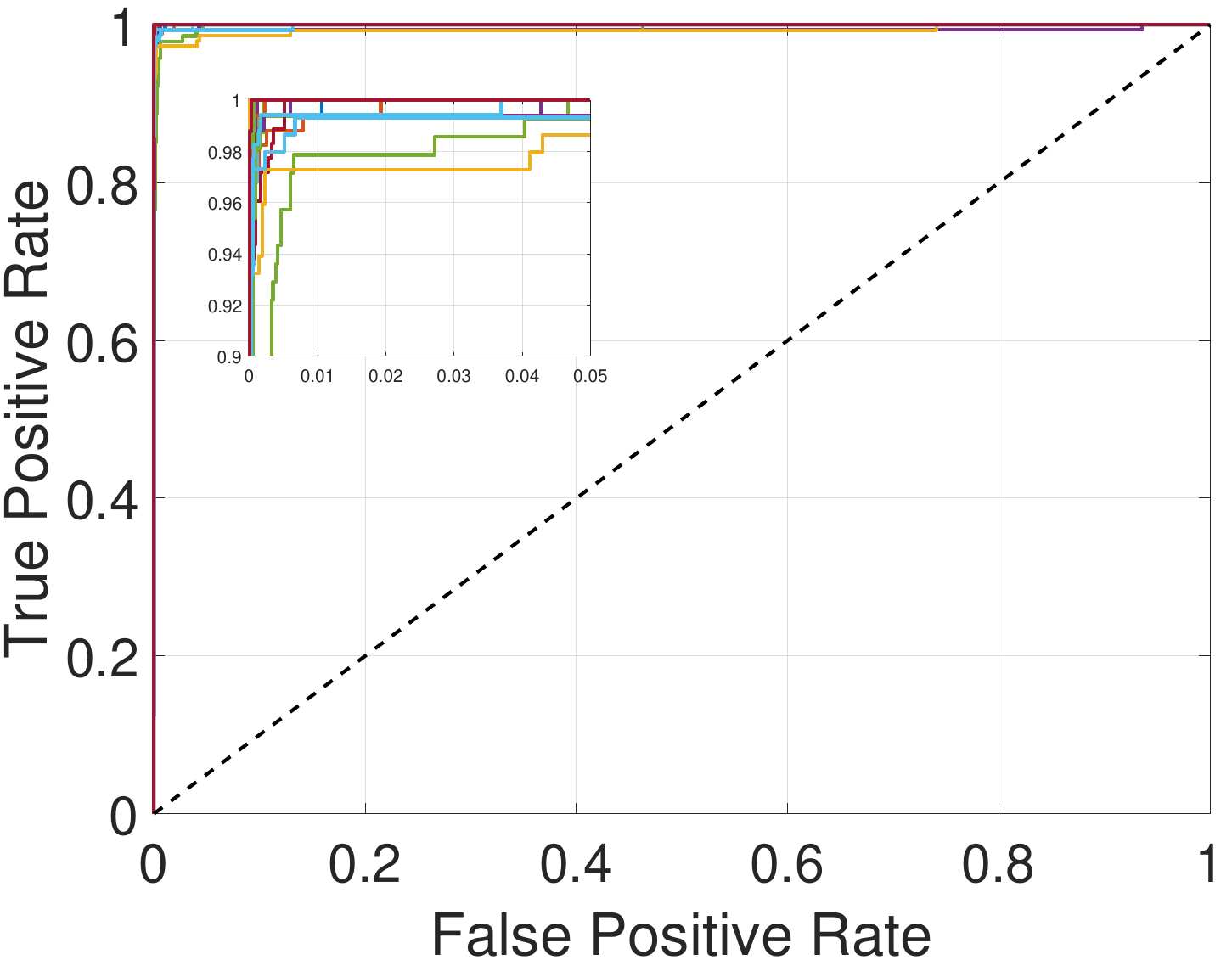}
    \caption{}
    \label{fig:roc_auc_transformar_tflite_quantized}
  \end{subfigure}
    \caption{ROC-AUC curves for the Transformer encoder model (a) TFLite and (b) TFLite Quantized, highlighting the model performance.}
    \label{fig:roc_auc}
\end{figure*}

\begin{table}[ht]
\centering
\caption{Training Time for CNN and Transformer Encoder models as a function of Batch Sizes}
    \resizebox{\columnwidth}{!}{%
        \begin{tabular}{|c|c|c|}
            \hline
            \textbf{Batch Size} & \textbf{CNN Training Time [s]} & \textbf{Transformer Training Time [s]} \\\hline
            1 & 3925.8 & 5092.2 \\\hline
            8 & 576.39 & 1794.3 \\\hline
            16 & 322.01 & 1380.9 \\\hline
            32 & 177.12 & 1173.1 \\\hline
            64 & 102.27 & 1128.8 \\ \hline
        \end{tabular}%
    }
\label{tab:batchsize_vs_time}
\end{table}

Figure~\ref{fig:tranning} depicts the training and validation accuracy and loss for both models.
In Figure~\ref{fig:cnn_train_validation}, the \ac{CNN} model training and validation accuracy starts to converge, with an accuracy $>$ 0.90, from the third epoch on. In the last epoch, the training accuracy is 0.99, while the validation accuracy is approximately 0.99. This shows that the model is able to learn by capturing the unique features of each transmitter in just a few training epochs. Similarly, the training and validation loss of the model decreases with increasing the number of epochs, illustrating that the model does not overfit. 

On the other hand, the Transformer Encoder model (Figure~\ref{fig:transformar_train_validation}) training accuracy increases more slowly with increasing the number of epochs; this demonstrates that the model takes a number of epochs to converge towards $>$ 0.90 accuracy. In the last epoch, the training accuracy was 0.97, while the validation accuracy was 0.98. Moreover, the training and validation loss decrease, implying that the model does not overfit.

\textbf{Evaluation} of the Keras trained and converted models, i.e., TFLite and Quantized TFLite, is done in two different setups: (i) using the raw IQ data with a defined sequence. and (ii) when randomizing the sequence. The second setup aims to assess the models' robustness against temporal changes and evaluate which model can capture the unique signal features and map them to its transmitter, even when altering their sequence. 

Figure~\ref{fig:confusion_matrix_keras_cnn} illustrates the confusion matrices when predicting the same sequence used in training (Figure~\ref{fig:confusion_matrix_keras_CNN}), and when randomizing the sequence (Figure~\ref{fig:confusion_matrix_keras_cnn_randomized}) for 
 the \ac{CNN} model. Similarly, Figure~\ref{fig:confusion_matrix_transformers} depicts the confusion matrices with the same aforementioned prediction setup for the Transformer Encoder, \ref{fig:confusion_matrix_keras_transformers} and \ref{fig:confusion_matrix_keras_transformers_randomized}, respectively. Unlike \ac{CNN}, the performance remains constant even when randomizing the IQ sequences. Table~\ref{tab:model_accuracies} summarizes the accuracies of the Keras trained, converted TFLite, and Quantized TFLite models.

 Furthermore, Figure~\ref{fig:roc_auc} shows the ROC-AUC for TFLite (Figure~\ref{fig:roc_auc_transformar_tflite}) and Quantized TFLite (Figure~\ref{fig:roc_auc_transformar_tflite_quantized}) Transformer Encoder models. For all 28 transmitters, both models retain very high AUC values ($>$ 0.90), indicating that both models are able to effectively distinguish between true positives and false positives across all the classes.

\definecolor{ao}{rgb}{0.0, 0.5, 0.0}
\begin{table}[!h]
\centering
\caption{Predication accuracy of the \ac{CNN}, Transformer, and Keras converted TFLite and Quantized models. }
\resizebox{\columnwidth}{!}{%
\begin{tabular}{|l|c|c|}
\hline
\textbf{Model Type} & \textbf{CNN Accuracy} & \textbf{Transformer Accuracy} \\ \hline
Trained Model & 0.99 & 0.98 \\\hline
TFLite & 0.99 & 0.98 \\\hline
Quantized TFLite & 0.99 & 0.98 \\ \hline\hline
TFLite Randomized & {\color{red}0.04} & {\color{ao}0.98} \\\hline
Quantized TFLite Randomized & {\color{red}0.04} & {\color{ao}0.98} \\ \hline
\end{tabular}%
}
\label{tab:model_accuracies}
\end{table}

\subsection{Inference on the Edge}
To evaluate the performance of the Keras converted and optimized models, we run 1000 inferences on a Raspberry Pi 4 and compute the average inference time, that is, the time to perform the prediction. \revised{The Raspberry Pi is equipped with 8GB of RAM}. Table~\ref{tab:inference} depicts the average inference times of the Keras converted \ac{CNN} and Transformer Encoder models, i.e., TFLite and Quantized TFLite. The \ac{CNN} model achieves faster inference times than the Transformer in both TFLite and Quantized TFLite. 

However, the Quantized model significantly reduces the inference time for both architectures, with the Transformer seeing more improvement. This is due to quantization, which reduces the numerical precision of the model's weights and activations from 32-bit floats to 8-bit integers. This reduces the model's size, makes it more memory-efficient, and speeds up processing, particularly on hardware optimized for integer arithmetic. Hence, Quantized models provide faster and more efficient inference without substantially compromising accuracy.

\begin{table}[!h]
\centering
\caption{Inference time for Transformer and \ac{CNN} models using TFLite and Quantized TFLite on Raspberry Pi 4.}
\label{tab:inference}
\resizebox{\columnwidth}{!}{%
\begin{tabular}{|c|c|c|}
\hline
\textbf{Model Type} & \textbf{Mean Inference Time [ms]} & \textbf{95\% Confidence Interval} \\ \hline
Transformer & 11.4334 & 0.0065 \\ \hline
CNN & 10.9049 & 0.0072 \\ \hline
Transformer Quantized & 0.5549 & 0.0005 \\ \hline
CNN Quantized & 0.3983 & 0.0010 \\ \hline
\end{tabular}%
}
\end{table}

\section{Related Work}
\label{sec:rw}
Enhancing security applications at the Edge is done through the deployment of \acf{AI}. It enables real-time attack detection~\cite{hussain2022jamming}, anomaly detection~\cite{8896997}, and automated recovery mechanisms~\cite{kok2022fogai}. \ac{AI}-based security solutions enable fast analysis of large amounts of data generated by \ac{IoT} devices to identify patterns indicative of attacks~\cite{zaman2021security}. \ac{TinyML}~\cite{ray2022review} enables the deployment of \ac{AI} models on resource-constrained \ac{IoT} and edge devices, allowing for the execution of \ac{ML} algorithms/models on such devices with limited computational power and memory. Thereby bringing advanced \ac{AI} capabilities closer to the data source and enhancing the overall security of \ac{IoT} networks.

Sun et al.~\cite{sun2023transformer} presented a transformer-based multi-feature extraction neural network for \ac{RFF} of different Bluetooth devices. First, the multi-scale features are extracted by a Transformer model, after which classification is performed by a ResNet module, resulting in an accuracy of 0.93. The use of the Transformer provided higher accuracy than other \ac{DL} methods. Feng et al.~\cite{feng2023lightweight} proposed a lightweight \ac{CNN} for \ac{RFF} on mobile devices. The model uses the IQ signal of the device as input, from which it extracts IQ-related and time-domain features. The performance of the model was tested on a public dataset with 16 USRP devices, achieving an accuracy of 0.85. Xie et al.~\cite{xie2018optimized} utilized coherent integration, multiresolution analysis, and Gaussian \ac{SVM} to optimize the classification of \ac{RF} fingerprints. The proposed optimized coherent integration algorithm works to de-noise and improve the \ac{SNR} of the received \ac{RF} fingerprint waveform. Then, the features of the waveform are extracted by waveform-based multiresolution analysis to reduce the dimension of the signals, which are classified by a classic \ac{SVM} algorithm.

\begin{table*}[!htbp]
\centering
\caption{The proposed implementation in comparison with existing ones. The first five references are the baseline solution, i.e., not tailored for Edge deployment. The following six contributions are tailored for Edge, and finally, the presented implementation in this work.}
\label{tab:compare}
\resizebox{0.8\textwidth}{!}{%
\begin{tabular}{|c|c|c|c|c|c|c|}
\hline
\textbf{Ref.} & \textbf{Architecture} & \textbf{Model Size} & \textbf{Number of Parameters} & \textbf{Model Accuracy} & \textbf{Tailored for Edge} \\ \hline
 \cite{wang2020convolutional} & CNN & - & - & 0.93 - 0.99 & \xmark \\\hline 
 \cite{wang2024radio} & CNN - ResNet18 & - & 11,200,000 & 0.95 - 0.98 & \xmark \\\hline 
 \cite{shenn2021radio} & CNN & 5.679 MB & 1,545,193 & 0.96 & \xmark \\\hline 
 \cite{wang2020radio} & CNN & - & - & 0.99 & \xmark \\\hline 
 \cite{sun2023transformer} & Transformer--ResNet & - & - & 0.93 & \xmark \\\hline 
 \hline
 \cite{feng2023lightweight} & CNN & - & 60,272 & 0.85 & \cmark \\\hline
 \cite{xie2018optimized} & SVM & - & - & 0.99 & \cmark \\\hline
 \cite{shen2021radio} & Transformer & - & 348,938 & 0.60 - 1.0 & \cmark \\ \hline
 \cite{jian2021radio} & CNN - ResNet50-1D & - & - & 0.60 - 0.70 & \cmark \\\hline
 \cite{wu2023radio} & CNN - ResNet50-1D (pruned) & - & - & 0.71 & \cmark \\ \hline
 \cite{shen2022towards} & CNN & 47.5 MB & 12,458,496 & 0.89 - 0.99 & \cmark \\ 
 \hline
 \hline
 \multirow{4}{*}{\textbf{This work}} 
  & CNN - TFLite & 462.02 KB & 116,808 & 0.99 & \cmark \\\cline{2-6} 
  & Transformer - TFLite & 210.20 KB & 47,964 & 0.98 & \cmark  \\\cline{2-6} 
  & CNN - Quantized TFlite & 123.80 KB & 116,808 & 0.99 & \cmark \\\cline{2-6} 
  & Transformer - Quantized TFlite & 73.27 KB & 47,964 & 0.98 & \cmark  \\ \hline

\end{tabular}%
}
\end{table*}

Shen et al.~\cite{shen2021radio} presented a transformer-based \ac{DL} model to classify signals from LoRA devices for \ac{RFF} in different \ac{SNR} conditions. The Transformer model is used to classify signals of variable length, which would not be possible with other \ac{DL} models such as \ac{CNN}. Only the packet preamble part of the LoRa packet is used for the classification. Data augmentation and multi-packet inference are also used to improve the models' performance in scenarios with low \ac{SNR}. Jian et al.~\cite{jian2021radio} proposed a structured pruning approach to train and sparsify \ac{RFF} \ac{DL} models to allow for their deployment on resource-constrained edge devices. Specifically, the convolutional layers of the \ac{CNN} are compressed, resulting in significantly faster inference time with a reduced number of parameters and an insignificant decrease in accuracy. The model is tested over several datasets that consist of IQ samples. Wu et al.~\cite{wu2023radio} presented a federated learning-based method to authenticate devices via \ac{RFF}. In addition to device authentication, the proposed method also addressed optimal resource allocation in order to reduce network delay when the computation is purely local, purely on the edge, or is hybrid local and edge. Shen et al.~\cite{shen2022towards} proposed a scalable deep metric learning approach for LoRa device authentication via \ac{RFF}. Initially, channel-independent spectrograms of known training devices are used to train an \ac{RF} fingerprint extractor that is able to generalize and output unique \ac{RF} fingerprints for different devices. The \ac{RF} fingerprints of legitimate devices in the network are then obtained from the \ac{RF} extractor and stored in a database, which is used by a k-NN algorithm to detect rogue devices and classify legitimate ones.

\section{Discussion}
\label{sec:discussion}
\textbf{Performance Evaluation and Model Effectiveness.} The performance evaluation (Section~\ref{sec:performance_evaluation}) shows that both the \ac{DL} and Transformer encoder models can be converted to perform \ac{RFF} on edge devices effectively. The Transformer encoder model, in particular, outperforms \ac{CNN} in terms of accuracy and robustness. This is due to the Transformer's multi-head self-attention mechanism, which captures dependencies and complex relationships within the IQ samples, better distinguishing subtle variations in the \ac{RF} fingerprints. Additionally, the Transformer's parallel processing capability enhances its understanding of the entire input sequence, while its flexible feature representation and regularization techniques reduce overfitting. This highlights the potential of Transformer architectures when deployed in edge computing environments, where computational resources are limited, yet high performance is required.

\textbf{Edge Device Deployment.} The deployment of \ac{AI} models on edge devices, such as Raspberry Pi~\cite{RaspberryPi}, presents several challenges, including limited processing power and memory constraints. Our implementation, supported by analysis, addresses these challenges by employing optimization methods provided by the TensorFlow library, namely model conversion to \ac{TFLite} and quantization. Such optimizations significantly reduce the model size and inference time, making it feasible to deploy \ac{CNN} or Transformer encoder for \ac{RFF} on edge devices such as the Raspberry Pi (and its variants), Espressif ESP32~\cite{esp32}, or Coral~\cite{Coral} without degradation in accuracy. This assumption is based on the specs of the aforementioned devices. 

\textbf{Comparison with the Baseline and Edge-tailored Solutions.} Table~\ref{tab:compare} compares the proposed solution against both (i) baseline and (ii) tailored for edge deployment. The baseline solutions, while they provide $>$ 0.90 model accuracy, are not tailored for deployment on the edge or resource-constrained devices. This is due to the model size, which is often in MB and, in some cases (based on the architecture and number of layers), larger than 100 MB. Hence, eliminating the feasibility of such deployment.

As for the Edge tailored solutions, (i) they rely on transfer learning (pre-trained models) that are not problem/domain-specific, i.e., not for \ac{RFF}. This results in increasing the number of models' parameters and, hence, its size (ii) their model architecture consists of multiple layers that affect the model learning ability, i.e., capturing the unique transmitter features, hence, a large number of parameters, a larger model size, and in some cases degradation in prediction accuracy.

It can be seen that our implementation offers significant advancements in several aspects. Notably, the trained models, including CNN TFLite and Transformer TFLite, achieve high accuracy rates of 0.99 and 0.98, respectively, higher than~\cite{jian2021radio}, or the highest accuracies reported in the literature, such as the SVM model~\cite{xie2018optimized}. Furthermore, our models are designed for edge deployment, as shown by their significantly smaller model sizes and parameter counts compared to the existing approaches. For instance, the CNN model presented in~\cite{shen2022towards} has a massive size of 47.5 MB and over 12 million parameters, while our CNN TFLite and Transformer TFLite models are only 462.02 KB and 210.20 KB, with 116,808 and 47,964 parameters, respectively. This substantial reduction in model size and complexity facilitates efficient inference on resource-constrained edge devices without compromising effectiveness. Additionally, our quantized TFlite models further enhance this efficiency, maintaining high accuracy while drastically reducing model size. The CNN-Quantized TFlite model maintains a high accuracy of 0.99 with a significantly reduced size of 123.80 KB, and the Transformer-Quantized TFlite model achieves an accuracy of 0.98 with an even smaller size of 73.27 KB. The TFlite quantized models offer a substantial reduction in model size, making them highly suitable for deployment on resource-constrained edge devices. This reduction in size is achieved without a significant drop in accuracy, demonstrating the robustness and efficiency of our proposed implementation.

\section{Conclusion and Future Work}
\label{sec:conclusions}
In this paper, we presented a novel methodology for deploying \ac{RFF} on edge devices to enhance the security of \ac{IoT} wireless networks by leveraging the unique device characteristics extracted from the raw IQ data. We implemented and evaluated two distinct \ac{DL} architectures--\ac{CNN} and Transformer encoder--optimized for edge deployment using TensorFlow Lite.

Our experimental results demonstrate that both models achieve an overwhelming accuracy ($>$ 0.95) and ROC-AUC score ($>$ 0.90) in identifying devices based on their \ac{RF} fingerprints, with the Transformer encoder sustaining performance in maintaining accuracy even when tested with samples that were not introduced during training. Additionally, we applied an optimization technique, specifically quantization, to ensure the models are lightweight and efficient, making them suitable for deployment on resource-constrained and edge devices. Furthermore, the models were evaluated on a Raspberry Pi, illustrating their feasibility for real-world edge deployment. Such lightweight implementations can be adopted in various applications that 5G and Beyond incorporate, e.g., \ac{IoD}, \ac{IoV}, and \ac{IoMT}, to name a few.

Future work will focus on expanding the scope of the dataset used to include a broader variety of IoT devices and diverse environmental conditions to validate the presented model's implementation generalizability. Additionally, we aim to explore other model architectures, fine-tune the hyperparameters, and apply advanced optimization techniques to enhance performance and efficiency further. We hope this contribution fulfills the security needs for Beyond 5G applications and paves the path towards deploying \ac{RFF} at the network edge. 

\balance
\bibliographystyle{IEEEtran}
\bibliography{main}

\begin{thebibliography}{10}
\providecommand{\url}[1]{#1}
\csname url@samestyle\endcsname
\providecommand{\newblock}{\relax}
\providecommand{\bibinfo}[2]{#2}
\providecommand{\BIBentrySTDinterwordspacing}{\spaceskip=0pt\relax}
\providecommand{\BIBentryALTinterwordstretchfactor}{4}
\providecommand{\BIBentryALTinterwordspacing}{\spaceskip=\fontdimen2\font plus
\BIBentryALTinterwordstretchfactor\fontdimen3\font minus
  \fontdimen4\font\relax}
\providecommand{\BIBforeignlanguage}[2]{{%
\expandafter\ifx\csname l@#1\endcsname\relax
\typeout{** WARNING: IEEEtran.bst: No hyphenation pattern has been}%
\typeout{** loaded for the language `#1'. Using the pattern for}%
\typeout{** the default language instead.}%
\else
\language=\csname l@#1\endcsname
\fi
#2}}
\providecommand{\BIBdecl}{\relax}
\BIBdecl

\bibitem{al2015internet}
A.~Al-Fuqaha, M.~Guizani, M.~Mohammadi, M.~Aledhari, and M.~Ayyash, ``Internet
  of things: A survey on enabling technologies, protocols, and applications,''
  \emph{IEEE communications surveys \& tutorials}, vol.~17, no.~4, pp.
  2347--2376, 2015.

\bibitem{narayanan2021harvestprint}
\BIBentryALTinterwordspacing
R.~Narayanan, A.~Varshney, and P.~Papadimitratos, ``{HarvestPrint: Securing
  Battery-Free Backscatter Tags through Fingerprinting},'' in \emph{ACM
  Workshop on Hot Topics in Networks (ACM HotNets)}, ser. HotNets '21.\hskip
  1em plus 0.5em minus 0.4em\relax New York, NY, USA: Association for Computing
  Machinery, November 2021, pp. 178 -- 184. [Online]. Available:
  \url{https://doi.org/10.1145/3484266.3487388}
\BIBentrySTDinterwordspacing

\bibitem{9170604}
P.~Tedeschi, S.~Sciancalepore, and R.~Di~Pietro, ``Security in energy
  harvesting networks: A survey of current solutions and research challenges,''
  \emph{IEEE Communications Surveys \& Tutorials}, vol.~22, no.~4, pp.
  2658--2693, 2020.

\bibitem{soltanieh2020review}
N.~Soltanieh, Y.~Norouzi, Y.~Yang, and N.~C. Karmakar, ``A review of radio
  frequency fingerprinting techniques,'' \emph{IEEE Journal of Radio Frequency
  Identification}, vol.~4, no.~3, pp. 222--233, 2020.

\bibitem{zhu2024secure}
C.~Zhu, K.~Li, J.~Hong, C.~Hua, and F.~Zou, ``A secure and private
  authentication based on radio frequency fingerprinting,'' in \emph{ICC
  2024-IEEE International Conference on Communications}.\hskip 1em plus 0.5em
  minus 0.4em\relax IEEE, 2024, pp. 2210--2215.

\bibitem{AlHazbiAHSSGOPP:C:2024}
S.~Al-Hazbi, A.~Hussain, S.~Sciancalepore, G.~Oligeri, and P.~Papadimitratos,
  ``Radio frequency fingerprinting via deep learning: Challenges and
  opportunities,'' in \emph{2024 International Wireless Communications and
  Mobile Computing (IWCMC)}, 2024, pp. 0824--0829.

\bibitem{jian2021radio}
T.~Jian, Y.~Gong, Z.~Zhan, R.~Shi, N.~Soltani, Z.~Wang, J.~Dy, K.~Chowdhury,
  Y.~Wang, and S.~Ioannidis, ``Radio frequency fingerprinting on the edge,''
  \emph{IEEE Transactions on Mobile Computing}, vol.~21, no.~11, pp.
  4078--4093, 2021.

\bibitem{wu2023radio}
W.~Wu, S.~Hu, D.~Lin, and T.~Yang, ``Radio-frequency fingerprinting for
  distributed iot networks: Authentication and qos optimization,'' \emph{IEEE
  Systems Journal}, vol.~17, no.~3, pp. 4440--4451, 2023.

\bibitem{he2016deep}
K.~He, X.~Zhang, S.~Ren, and J.~Sun, ``Deep residual learning for image
  recognition,'' in \emph{Proceedings of the IEEE conference on computer vision
  and pattern recognition}, 2016, pp. 770--778.

\bibitem{hanna_wisig_2022}
S.~Hanna, S.~Karunaratne, and D.~Cabric, ``Wisig: A large-scale wifi signal
  dataset for receiver and channel agnostic rf fingerprinting,'' \emph{IEEE
  Access}, vol.~10, p. 22808–22818, 2022.

\bibitem{alzubaidi2021review}
L.~Alzubaidi, J.~Zhang, A.~J. Humaidi, A.~Al-Dujaili, Y.~Duan, O.~Al-Shamma,
  J.~Santamar{\'\i}a, M.~A. Fadhel, M.~Al-Amidie, and L.~Farhan, ``Review of
  deep learning: concepts, cnn architectures, challenges, applications, future
  directions,'' \emph{Journal of big Data}, vol.~8, pp. 1--74, 2021.

\bibitem{o2015introduction}
K.~O'shea and R.~Nash, ``An introduction to convolutional neural networks,''
  \emph{arXiv preprint arXiv:1511.08458}, 2015.

\bibitem{vaswani2017attention}
A.~Vaswani, N.~Shazeer, N.~Parmar, J.~Uszkoreit, L.~Jones, A.~N. Gomez,
  {\L}.~Kaiser, and I.~Polosukhin, ``Attention is all you need,''
  \emph{Advances in neural information processing systems}, vol.~30, 2017.

\bibitem{hayyolalam2021edge}
V.~Hayyolalam, M.~Aloqaily, {\"O}.~{\"O}zkasap, and M.~Guizani, ``Edge
  intelligence for empowering iot-based healthcare systems,'' \emph{IEEE
  Wireless Communications}, vol.~28, no.~3, pp. 6--14, 2021.

\bibitem{yan2022survey}
W.~Yan, Z.~Wang, H.~Wang, W.~Wang, J.~Li, and X.~Gui, ``Survey on recent smart
  gateways for smart home: Systems, technologies, and challenges,''
  \emph{Transactions on Emerging Telecommunications Technologies}, vol.~33,
  no.~6, p. e4067, 2022.

\bibitem{paiva2021enabling}
S.~Paiva, M.~A. Ahad, G.~Tripathi, N.~Feroz, and G.~Casalino, ``Enabling
  technologies for urban smart mobility: Recent trends, opportunities and
  challenges,'' \emph{Sensors}, vol.~21, no.~6, p. 2143, 2021.

\bibitem{singh2023edge}
R.~Singh and S.~S. Gill, ``Edge ai: a survey,'' \emph{Internet of Things and
  Cyber-Physical Systems}, vol.~3, pp. 71--92, 2023.

\bibitem{liang2021pruning}
T.~Liang, J.~Glossner, L.~Wang, S.~Shi, and X.~Zhang, ``Pruning and
  quantization for deep neural network acceleration: A survey,''
  \emph{Neurocomputing}, vol. 461, pp. 370--403, 2021.

\bibitem{elmaghbubdistinguishable}
A.~Elmaghbub and B.~Hamdaoui, ``Distinguishable iq feature representation for
  domain-adaptation learning of wifi device fingerprints,'' \emph{IEEE
  Transactions on Machine Learning in Communications and Networking}, pp.
  1404--1423, August 2024.

\bibitem{10288752}
------, ``A needle in a haystack: Distinguishable deep neural network features
  for domain-agnostic device fingerprinting,'' in \emph{2023 IEEE Conference on
  Communications and Network Security (CNS)}, 2023, pp. 1--9.

\bibitem{9040673}
C.~Geng, S.-J. Huang, and S.~Chen, ``Recent advances in open set recognition: A
  survey,'' \emph{IEEE Transactions on Pattern Analysis and Machine
  Intelligence}, vol.~43, no.~10, pp. 3614--3631, 2021.

\bibitem{WANG2024124537}
\BIBentryALTinterwordspacing
X.~Wang, Q.~Wang, L.~Fang, M.~Hua, Y.~Jiang, and Y.~Hu, ``Radio frequency
  fingerprint authentication based on feature fusion and contrastive
  learning,'' \emph{Expert Systems with Applications}, vol. 255, p. 124537,
  2024. [Online]. Available:
  \url{https://www.sciencedirect.com/science/article/pii/S0957417424014040}
\BIBentrySTDinterwordspacing

\bibitem{tf}
\BIBentryALTinterwordspacing
{Google}. (2024) {TensorFlow}. Accessed: Jun 2024. [Online]. Available:
  \url{https://www.tensorflow.org}
\BIBentrySTDinterwordspacing

\bibitem{keras}
\BIBentryALTinterwordspacing
{François Chollet}. (2024) {Keras}. Accessed: Jun 2024. [Online]. Available:
  \url{https://keras.io/}
\BIBentrySTDinterwordspacing

\bibitem{kingma2014adam}
D.~P. Kingma and J.~Ba, ``Adam: A method for stochastic optimization,''
  \emph{arXiv preprint arXiv:1412.6980}, 2014.

\bibitem{adam}
\BIBentryALTinterwordspacing
{Keras}. (2024) {Adam Optimizer}. Accessed: Jun 2024. [Online]. Available:
  \url{https://keras.io/api/optimizers/adam/}
\BIBentrySTDinterwordspacing

\bibitem{hussain2022jamming}
\BIBentryALTinterwordspacing
A.~Hussain, N.~Abughanam, J.~Qadir, and A.~Mohamed, ``Jamming detection in iot
  wireless networks: An edge-ai based approach,'' in \emph{Proceedings of the
  12th International Conference on the Internet of Things}, ser. IoT '22.\hskip
  1em plus 0.5em minus 0.4em\relax New York, NY, USA: Association for Computing
  Machinery, 2023, p. 57–64. [Online]. Available:
  \url{https://doi.org/10.1145/3567445.3567456}
\BIBentrySTDinterwordspacing

\bibitem{tflite_optimize}
\BIBentryALTinterwordspacing
{Google}. (2024) {TFLite Optimize}. Accessed: Jun 2024. [Online]. Available:
  \url{https://www.tensorflow.org/api_docs/python/tf/lite/Optimize}
\BIBentrySTDinterwordspacing

\bibitem{8896997}
B.~Hussain, Q.~Du, A.~Imran, and M.~A. Imran, ``Artificial intelligence-powered
  mobile edge computing-based anomaly detection in cellular networks,''
  \emph{IEEE Transactions on Industrial Informatics}, vol.~16, no.~8, pp.
  4986--4996, 2020.

\bibitem{kok2022fogai}
{\.I}.~K{\"o}k, F.~Y. Okay, and S.~{\"O}zdemir, ``Fogai: An ai-supported fog
  controller for next generation iot,'' \emph{Internet of Things}, vol.~19, p.
  100572, 2022.

\bibitem{zaman2021security}
S.~Zaman, K.~Alhazmi, M.~A. Aseeri, M.~R. Ahmed, R.~T. Khan, M.~S. Kaiser, and
  M.~Mahmud, ``Security threats and artificial intelligence based
  countermeasures for internet of things networks: A comprehensive survey,''
  \emph{IEEE Access}, vol.~9, pp. 94\,668--94\,690, 2021.

\bibitem{ray2022review}
P.~P. Ray, ``A review on tinyml: State-of-the-art and prospects,''
  \emph{Journal of King Saud University-Computer and Information Sciences},
  vol.~34, no.~4, pp. 1595--1623, 2022.

\bibitem{sun2023transformer}
C.~Sun, X.~Chen, W.~Wang, and H.~Yin, ``A transformer-based multi-feature
  extraction neural network for bluetooth devices identification,'' in
  \emph{2023 9th International Conference on Computer and Communications
  (ICCC)}.\hskip 1em plus 0.5em minus 0.4em\relax IEEE, 2023, pp. 1469--1473.

\bibitem{feng2023lightweight}
J.~Feng, X.~Tang, B.~Zhang, and Y.~Ren, ``Lightweight cnn-based rf fingerprint
  recognition method,'' in \emph{2023 8th International Conference on Computer
  and Communication Systems (ICCCS)}.\hskip 1em plus 0.5em minus 0.4em\relax
  IEEE, 2023, pp. 1031--1035.

\bibitem{xie2018optimized}
F.~Xie, H.~Wen, Y.~Li, S.~Chen, L.~Hu, Y.~Chen, and H.~Song, ``Optimized
  coherent integration-based radio frequency fingerprinting in internet of
  things,'' \emph{IEEE Internet of Things Journal}, vol.~5, no.~5, pp.
  3967--3977, 2018.

\bibitem{wang2020convolutional}
S.~Wang, L.~Peng, H.~Fu, A.~Hu, and X.~Zhou, ``A convolutional neural
  network-based rf fingerprinting identification scheme for mobile phones,'' in
  \emph{IEEE INFOCOM 2020-IEEE Conference on Computer Communications Workshops
  (INFOCOM WKSHPS)}.\hskip 1em plus 0.5em minus 0.4em\relax IEEE, 2020, pp.
  115--120.

\bibitem{wang2024radio}
X.~Wang, Q.~Wang, L.~Fang, M.~Hua, Y.~Jiang, and Y.~Hu, ``Radio frequency
  fingerprint authentication based on feature fusion and contrastive
  learning,'' \emph{Expert Systems with Applications}, vol. 255, p. 124537,
  2024.

\bibitem{shenn2021radio}
G.~Shen, J.~Zhang, A.~Marshall, L.~Peng, and X.~Wang, ``Radio frequency
  fingerprint identification for lora using deep learning,'' \emph{IEEE Journal
  on Selected Areas in Communications}, vol.~39, no.~8, pp. 2604--2616, 2021.

\bibitem{wang2020radio}
S.~Wang, H.~Jiang, X.~Fang, Y.~Ying, J.~Li, and B.~Zhang, ``Radio frequency
  fingerprint identification based on deep complex residual network,''
  \emph{IEEE access}, vol.~8, pp. 204\,417--204\,424, 2020.

\bibitem{shen2021radio}
G.~Shen, J.~Zhang, A.~Marshall, M.~Valkama, and J.~Cavallaro, ``Radio frequency
  fingerprint identification for security in low-cost iot devices,'' in
  \emph{2021 55th Asilomar conference on signals, systems, and
  computers}.\hskip 1em plus 0.5em minus 0.4em\relax IEEE, 2021, pp. 309--313.

\bibitem{shen2022towards}
G.~Shen, J.~Zhang, A.~Marshall, and J.~R. Cavallaro, ``Towards scalable and
  channel-robust radio frequency fingerprint identification for lora,''
  \emph{IEEE Transactions on Information Forensics and Security}, vol.~17, pp.
  774--787, 2022.

\bibitem{RaspberryPi}
\BIBentryALTinterwordspacing
{Raspberry Pi}. {Raspberry Pi 4 Model B}. Accessed April 2023. [Online].
  Available: \url{https://www.raspberrypi.com/products/raspberry-pi-4-model-b/}
\BIBentrySTDinterwordspacing

\bibitem{esp32}
\BIBentryALTinterwordspacing
{Espressif}. {Espressif ESP32 DevKitC}. Accessed: Jun 2024. [Online].
  Available:
  \url{https://www.espressif.com/en/products/devkits/esp32-devkitc/overview}
\BIBentrySTDinterwordspacing

\bibitem{Coral}
\BIBentryALTinterwordspacing
{Coral}. (2024) {Coral}. Accessed: Jun 2024. [Online]. Available:
  \url{https://coral.ai/}
\BIBentrySTDinterwordspacing

\end{thebibliography}

\newpage
\begin{IEEEbiographynophoto}{Ahmed Mohamed Hussain}
is currently a Predoctoral Researcher working with the Networked Systems Security (NSS) Group at KTH Royal Institute of Technology. Prior to joining KTH, he worked as a researcher in the Computer Science and Engineering department at Qatar University, Doha, Qatar. He received his Master's in Computer Science at Uppsala University, Uppsala, Sweden. His research interests include IoT, security and privacy, networks (particularly wireless) security, Artificial Intelligence for Cybersecurity, and Physical Layer Security.

\end{IEEEbiographynophoto}
\vspace{-33pt}
\begin{IEEEbiographynophoto}{Nada Abughanam} is currently a Research Assistant in the Electrical Engineering department at Qatar University, Doha, Qatar. She holds a Bachelor's in Electrical Engineering and a Master's in Computer Engineering. Her research interests include human-computer interaction, Artificial intelligence, and IoT.

\end{IEEEbiographynophoto}
\vspace{-33pt}
\begin{IEEEbiographynophoto}{Panos Papadimitratos} earned his Ph.D. degree from Cornell University, Ithaca, NY, USA. He leads the Networked Systems Security group at KTH, Stockholm, Sweden. He serves as a member (and currently chair) of the ACM WiSec conference steering committee, member of the PETS Advisory Board, and the CANS conference steering committee. Panos is an IEEE Fellow, a Young Academy of Europe Fellow, a Knut and Alice Wallenberg Academy Fellow, and an ACM Distinguished Member. His group webpage is: www.eecs.kth.se/nss.
\end{IEEEbiographynophoto}

\vfill

\end{document}